\documentclass[10pt,journal,compsoc]{IEEEtran}

\usepackage{tabularx} 
\usepackage{svg}
\usepackage{adjustbox}

\ifCLASSOPTIONcompsoc
  \usepackage[nocompress]{cite}
\else
  \usepackage{cite}
\fi

\ifCLASSINFOpdf
\else
\fi

\usepackage{amsmath}

\usepackage{array}

\usepackage{amsmath}
\usepackage{amssymb}
\usepackage{amsthm}
\usepackage{mdwmath}
\usepackage{mdwtab}
\usepackage{mathtools}
\DeclareMathOperator{\Tr}{Tr}

 \newtheorem{lemma} {Lemma}

 \newtheorem{definition}{Definition} 
 \newtheorem{remark}{Remark} 
\usepackage{booktabs}
\usepackage{amsmath}

\usepackage{amssymb}
\usepackage{amsthm}
\usepackage{mdwmath}
\usepackage{mdwtab}
\usepackage{mathtools}

\usepackage{pgfplots}
\pgfplotsset{compat=newest}
\usepackage{mathrsfs}
\usetikzlibrary{arrows}

\usepackage{fontawesome5}

\ifCLASSOPTIONcompsoc
\usepackage[caption=false,font=footnotesize,labelfont=sf,textfont=sf]{subfig}
\else
\usepackage[caption=false,font=footnotesize]{subfig}
\fi

\usepackage{url}

\usepackage{bm}
	
\usepackage[ruled,linesnumbered]{algorithm2e}

\DeclareMathOperator{\diag}{diag} 
	
\usepackage{hyperref} 

\usepackage{graphbox,graphicx}

\usepackage{tcolorbox}

\ifCLASSOPTIONcompsoc
  \usepackage[caption=false,font=footnotesize,labelfont=sf,textfont=sf]{subfig}
\else
  \usepackage[caption=false,font=footnotesize]{subfig}
\fi

\usepackage{pgfplots}
\DeclareUnicodeCharacter{2212}{−}
\usepgfplotslibrary{groupplots,dateplot}
\usetikzlibrary{patterns,shapes.arrows}
\pgfplotsset{compat=newest}

\usepackage{booktabs}
\usepackage{multirow}
\usepackage{makecell}
 \usepackage{graphicx}

\usepackage{url}
\usepackage{bm}

\usepackage[ruled,linesnumbered]{algorithm2e}

\usepackage{hyperref}  

\begin{document}
%\title{Robust Pose Estimation by Rotation Voting}
\title{Linearly Solving Robust Rotation Estimation}

\author{Yinlong~Liu, Tianyu~Huang
        and~Zhi-Xin Yang$^\sharp$
        % <-this % stops a space
\IEEEcompsocitemizethanks{\IEEEcompsocthanksitem Yinlong Liu and Zhi-Xin Yang are with State Key Laboratory of Internet of Things for Smart City (SKL-IOTSC), University of Macau, Macau.\protect\\
E-mail: \{yinlongliu,  zxyang\}@um.edu.mo
\IEEEcompsocthanksitem  Tianyu Huang is with the Hong Kong Centre For Logistics Robotics, The Chinese University of Hong Kong, Hong Kong. \protect\\ 
E-mail: tianyuhuang@cuhk.edu.hk
\IEEEcompsocthanksitem $^\sharp$Zhi-Xin Yang is the corresponding author.
}% <-this % stops a space

\thanks{Manuscript received April 19, 20XX; revised August 26, 20XX.}}

\markboth{Journal of \LaTeX\ Class Files,~Vol.~14, No.~8, August~2025}%
{Shell \MakeLowercase{\textit{et al.}}: Bare Advanced Demo of IEEEtran.cls for IEEE Computer Society Journals}



\IEEEtitleabstractindextext{%
\begin{abstract}
Rotation estimation plays a fundamental role in computer vision and robot tasks, and extremely robust rotation estimation is significantly useful for safety-critical applications. Typically, estimating a rotation is considered a non-linear and non-convex optimization problem that requires careful design. However, in this paper, we provide some new perspectives that solving a rotation estimation problem can be reformulated as solving a linear model fitting problem without dropping any constraints and without introducing any singularities. In addition, we explore the dual structure of a rotation motion, revealing that it can be represented as a great circle on a quaternion sphere surface.  Accordingly, we propose an easily understandable voting-based method to solve rotation estimation. The proposed method exhibits exceptional robustness to noise and outliers and can be computed in parallel with graphics processing units (GPUs) effortlessly. Particularly, leveraging the power of GPUs, the proposed method can obtain a satisfactory rotation solution for large-scale ($10 ^6 $) and severely corrupted  (99$\%$ outlier ratio) rotation estimation problems under 0.5 seconds. Furthermore, to validate our theoretical framework and demonstrate the superiority of our proposed method, we conduct controlled experiments and real-world dataset experiments. These experiments provide compelling evidence supporting the effectiveness and robustness of our approach in solving rotation estimation problems.
\end{abstract}

}

\maketitle

\IEEEdisplaynontitleabstractindextext

% \fi
\IEEEpeerreviewmaketitle

\ifCLASSOPTIONcompsoc
\IEEEraisesectionheading{\section{Introduction}\label{sec:introduction}}
\else
\section{Introduction}
\label{sec:introduction}
\fi

\IEEEPARstart{R}{otation} estimation is a fundamental problem in the fields of computer vision and robotics~\cite{bustos2016fast}. It is a core part of rigid pose estimation, and many intelligent applications rely on its accurate and reliable solution, e.g., auto-driving and robot grasping~\cite{camposeco2018hybrid,szeliski2022computer}. 
In addition, it is also widely used in aerospace engineering. For example, it is equivalently related to Wahba's problem~\cite{markley2006love,shuster2006generalized}, which is to estimate spacecraft attitude~\cite{markley2013equivalence,psiaki2012numerical} and has a more than half-century history~\cite{markley199930}. 
Furthermore,the star identification task~\cite{zhang2016star} can be also formulated as solving a rotation estimation problem~\cite{chng2023rosia}.

Mathematically, given $N$ observations $\left\{\bm{x}_i,\bm{y}_i\right\}_{i=1}^N$, the target of rotation estimation is to find a rotation $\mathbf{R}\in \mathbb{SO}(3)$ that satisfies the following motions as much as possible,
\begin{equation}
	\mathbf{R}\bm{x}_i=\bm{y}_i,\quad i=1\cdots N
\end{equation}
In real applications, it is almost impossible to obtain perfect observations to calculate the accurate rotation, and it is unavoidable to have noise and outliers in the inputs~\cite{yang2023certifiably,chin2022maximum}. Therefore, there is a significant demand to design robust algorithms that can obtain a satisfactory rotation estimation result even with corrupted inputs~\cite{antonante2021outlier}. Particularly, an extremely robust rotation estimation algorithm is strongly needed for safety-critical applications, such as self-driving systems~\cite{marti2019review} and medical robots~\cite{bianchi2019localization,liu20182d}.

In this paper, our main aim is to provide an extremely robust and highly efficient rotation estimation method. Specifically, the provided method should obtain satisfactory rotation motion even if the inputs are contaminated by noise and outliers (i.e., mismatches). Furthermore, the rotation estimation method should be efficient enough to be a fundamental ingredient in solving subsequent tasks.

\begin{figure}
	\begin{tabular}{cc}
		\includegraphics[width=0.48\linewidth]{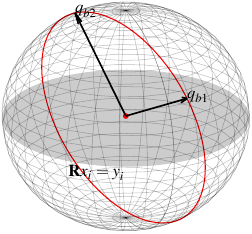}
		&
		\includegraphics[width=0.48\linewidth]{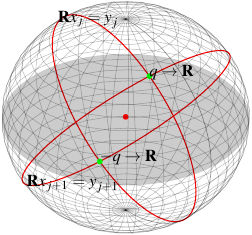}
		\\
		(a) &(b)
	\end{tabular}
	\caption{The visualization of quaternion circle. For better understanding, we draw the quaternion circles in $\mathbb{R}^3$. (a) Given $\mathbf{R}\bm{x}_i=\bm{y}_i$, the solution of $\mathbf{R}$ must be in a great circle (one-dimensional structure) in the unit quaternion sphere. (b) Given two different input observations, $\mathbf{R}$ can be solved. Geometrically, the to-be-solved rotation is the intersection of two quaternion circles, which are obtained from the input observations. Specifically, there are two symmetrical intersecting points in the unit quaternion sphere surface, i.e., $\bm{q}$ and $-\bm{q}$. They are however corresponding to the same rotation.  }
	\label{fig:great_circle}
\end{figure}
Methodologically, at first glance, outlier-robust rotation motion estimation is a non-linear~\cite{yang2019quaternion} and non-convex~\cite{olsson2009branch} model fitting problem~\cite{chin2022maximum} due to $\mathbf{R}\in\mathbb{SO}(3)$, which is a non-linear and non-convex solution domain~\cite{iglesias2020global}.  To obtain a satisfactory rotation, a sophisticated and special algorithm should be designed~\cite{parra2016robust,bazin2013globally}.  However, we argue that given a rotation motion, it can be reformulated as a linear system that includes two equations without dropping any constraints~\cite{carlone2018convex} and without adding any singularities~\cite{mortari2007optimal}. Accordingly, solving rotation estimation can be considered as solving a linear system. Therefore,  some robust linear model fitting methods, which have excellent performance in accuracy and robustness~\cite{chin2022maximum,liu2022globallya}, can also be applied to solve robust rotation estimation. Consequently, we design a rotation voting method to solve the robust rotation estimation problem, and the method is inspired by Hough transformation~\cite{hart2009hough,mukhopadhyay2015survey}

The rest of this paper is organized as follows. Section 2 presents a brief review of related rotation estimation methods with/without outlier inputs. Section 3 describes the motivation and contributions of our work. Furthermore, Section 4 provides details of our theory and rotation voting method for the outlier-robust rotation estimation problem. Section 5 illustrates the experiments, and Section 6 concludes this paper.
\section{Related Work}
According to whether the solution can obtain satisfactory solutions from outlier-contaminated data, related works are divided into two categories~\cite{parra2016robust}: (1) outlier-free algorithms and (2) outlier-robust algorithms.

\subsection{Outlier-Free Solutions}
Given $N$ observations $\left\{\bm{x}_i,\bm{y}_i\right\}_{i=1}^N$, without considering outliers (mismatches), the objective is typically formulated as a least squares problem,
\begin{equation}
	\min_{\mathbf{R}\in\mathbb{SO}(3)} \sum_{i=1}^{N}\|\mathbf{R}\boldsymbol{x}_i-\boldsymbol{y}_i\|^2 \label{eq:objective}
\end{equation}
where $\|\cdot\|$ is the euclidean norm. 
There are numerous methods to solve this optimization problem~\cite{lourakis2018efficient,markley199930}. At least two closed-form solutions should be mentioned~\cite{eggert1997estimating}. Specifically, the objective can be reformulated as 
\begin{equation}
	\max_{\mathbf{R}\in\mathbb{SO}(3)}\sum_{i=1}^{N}\bm{y}_i^T\mathbf{R}\bm{x}_i\label{eq:objective-simple}
\end{equation}

(1) \textbf{SVD (singular value decomposition) based algorithm}~\cite{arun1987least,umeyama1991least}. First, construct an attitude profile matrix and decompose it using SVD,
\begin{equation}
	\mathbf{B}=\sum_{i=1}^{N}\bm{y}_i\bm{x}_i^T,\text{and}, \mathbf{B}=\mathbf{U}\mathbf{\Sigma}\mathbf{V}^T
\end{equation}
where $\mathbf{B}\in\mathbb{R}^{3\times3}$ is called attitude profile matrix~\cite{markley2013equivalence}; $\mathbf{U}\mathbf{U}^T=\mathbf{I}$ and $\mathbf{V}\mathbf{V}^T=\mathbf{I}$; $\mathbf{\Sigma}=\diag\left(\begin{bmatrix}
	\Sigma_0&\Sigma_1&\Sigma_2
\end{bmatrix}\right)$ is singular values matrix and $	\Sigma_0\geq\Sigma_1\geq\Sigma_2\geq0$. Accordingly, Eq.~\eqref{eq:objective-simple} can be reformulated as
\begin{align}
	&\max \langle\mathbf{R},\mathbf{B}\rangle=\max \langle\mathbf{R},\mathbf{U}\mathbf{\Sigma}\mathbf{V}^T\rangle=\max \langle\mathbf{U}^T\mathbf{R}\mathbf{V},\mathbf{\Sigma}\rangle
	\\
&	=\max \langle\mathbf{S},\mathbf{\Sigma}\rangle=
	\max \Sigma_0S_0+\Sigma_1S_1+\Sigma_2S_2
\end{align}
where $\langle\cdot,\cdot\rangle$ is to calculate Frobenius inner product of two matrix; $\mathbf{S}=\mathbf{U}^T\mathbf{R}\mathbf{V}$ is an orthogonal matrix;  $\Sigma_0,\Sigma_1,\Sigma_2$ are diagonal elements of $\mathbf{S}$ and their absolute values are all not greater than 1.  Therefore, when $\mathbf{S}=\mathbf{I}$, Eq.~\eqref{eq:objective-simple} gets maximum. In addition, to ensure $\det\left(\mathbf{R}\right)=1$, the optimal rotation should be
\begin{equation}
	\mathbf{R}^*=\mathbf{U}
	\diag\left(
	\begin{bmatrix}
		1&1&\det\left(\mathbf{U}\mathbf{V}^T\right)
	\end{bmatrix}
	\right)
	\mathbf{V}^T
\end{equation}
%where $\diag(\cdot)$ is to construct a diagonal matrix, and $\det(\cdot)$ is determinant of a square matrix

(2) \textbf{Quaternion based algorithm}~\cite{horn1987closed,bayro2021survey}. It is well-known the rotation matrix $\mathbf{R}$ can be represented by unit quaternion $\bm{q}$, and Eq.~\eqref{eq:objective-simple} can be reformulated as~\cite{horn1987closed}, 
\begin{equation}
	\max_{\bm{q}\in\mathbb{S}^3}\bm{q}^T\mathbf{M}\bm{q}
	\label{eq:quat-objective}
\end{equation}
where $\mathbf{M}\in\mathbb{R}^{4\times4}$ is a symmetric matrix and is constructed from all observations (see appendix).
Therefore, it can be decomposed as
\begin{equation}
	\mathbf{M}=\mathbf{O}\mathbf{\Lambda}\mathbf{O}^T
\end{equation}
where $\mathbf{O}\in\mathbb{R}^{4\times4}$ and 	$\mathbf{O}$  is an orthogonal matrix $\mathbf{O}\mathbf{O}^T=\mathbf{I}$; $\mathbf{\Lambda}=\diag\begin{bmatrix}
	\lambda_0&\lambda_1&\lambda_2&\lambda_3
\end{bmatrix}$ and $\lambda_0\geq\lambda_1\geq\lambda_2\geq\lambda_3$ are eigenvalue of $\mathbf{M}$.
 Let $\bm{\Gamma}=\left[\Gamma_0,\Gamma_1,\Gamma_2,\Gamma_3\right]^T=\mathbf{O}^T\bm{q}$. Then 
\begin{align}
	\bm{\Gamma}^T\bm{\Gamma}= \bm{q}^T\mathbf{O}\mathbf{O}^T\bm{q}&=\Gamma_0^2+\Gamma_1^2+\Gamma_2^2+\Gamma_3^2=1
\\
	\bm{q}^T\mathbf{M}\bm{q}=\bm{\Gamma}^T\Lambda\bm{\Gamma}&=\lambda_0\Gamma_0^2+\lambda_1\Gamma_1^2+\lambda_2\Gamma_2^2+\lambda_3\Gamma_3^2
\end{align}
Therefore, Eq.~\eqref{eq:quat-objective} achieves its maximum when $\Gamma_0=1$ and $\Gamma_1=\Gamma_2=\Gamma_3=0$. 
\begin{equation}
\max \bm{q}^T\mathbf{M}\bm{q}=\lambda_0,\text{and},\bm{q}^*=\mathbf{O}\begin{bmatrix}
	1&0&0&0
\end{bmatrix}^T
\end{equation}
In other words, the rotation $\bm{q}$ that maximizes Eq.~\eqref{eq:quat-objective} is the eigenvector that corresponds to the largest eigenvalue of $\mathbf{M}$.

In addition to closed-form solutions, there are many iterative ways~\cite{kruzhilov2014iteration,peng2022arcs} to solve the rotation optimization problem. However, in terms of efficiency and robustness, closed-form solutions are generally superior to iterative methods, as iterative solutions frequently suffer from a lack of convergence guarantee, becoming
trapped in local minima, and requiring a good starting estimate~\cite{eggert1997estimating}.

\subsection{Outlier-Robust Solutions}
When the inputs are contaminated by outliers, the results of least squares-based methods may be significantly biased~\cite{chin2022maximum}. To address this problem, outlier-robust rotation algorithms are developed.

RANSAC (RANdom SAmple Consensus)~\cite{fischler1981random} algorithm is the de facto standard outlier robust solution in the computer vision field~\cite{chin2022maximum}, which repeatedly and randomly samples minimal or non-minimal observations and calculates solution candidates~\cite{DBLP:conf/bmvc/FragosoSST17}. The best of them will be returned as the final one. Typically, RANSAC is fast and robust on many occasions. However, some researchers argue that RASNAC cannot guarantee the quality of the returned solution, due to its inherent randomness~\cite{antonante2021outlier,chin2022maximum,liu2022globallya}. 
Furthermore, M-estimators~\cite{andersen2008modern} are applied to formulate the objective function to suppress outliers and robustly obtain the maximum likelihood estimation. However, the objective loss is typically non-convex to reduce the impact of outliers, and it usually leads to non-convex optimization~\cite{chin2022maximum}.  Due to the hardness~\cite{antonante2021outlier} of seeking the global optimum of a non-convex problem~\cite{carlone2023estimation}, it is not easy to find a globally optimal rotation estimation~\cite{parra2016robust}. To alleviate this problem, many researchers try to formulate a relaxed version of the objective and provide provable uncertainty bounds for optimized solutions, which are called certifiable solutions~\cite{yang2023certifiably,barfoot2023certifiably,yang2019quaternion}. 

To exactly solve the outlier-robust rotation estimation, a direct idea is to remove all outliers and use only inliers to solve the outlier-free rotation estimation~\cite{parra2018guaranteed,olsson2010outlier,Huang-RAL2024}. Alternatively, it can be considered to be using a 0-1 binary loss function to distinguish which input observations are inliers and which are outliers~\cite{chin2022maximum}. However,  it is theoretically shown that removing outliers is just as difficult as the original estimation problem~\cite{antonante2021outlier}. Fortunately, rotation motion estimation is a geometrical problem~\cite{hartley2003multiple}, therefore, some special geometrical constraints can be applied to remove outliers~\cite{parra2018guaranteed}. Furthermore, to avoid incorrect solutions and provide satisfactory rotation with optimality guarantees, considering rotational geometrical constraints~\cite{hartley2009global}, rotation search theory is developed~\cite{hartley2009global,li20073d,liu20182d,parra2016robust,bazin2013globally,campbell2020globally, Huang-TPAMI2024}. It applies the branch-and-bound framework to systematically search the entire $\mathbb{SO}(3)$ domain to obtain the optimal rotation, which theoretically can return a global optimum~\cite{li20073d}. Therefore, it is particularly suitable for safety-critical applications~\cite{liu20182d}, while this type of method is time-consuming~\cite{yang2023certifiably, Huang-CVPR2024}.

To accelerate the globally optimal methods, recently, pairwise constraints are widely applied to reduce the difficulty of solving rigid pose estimation, such as translation and rotation invariant measurements~ \cite{yang2020teaser,parra2016robust,liu2018efficient,li2023qgore,li2021practical}. What is particularly worth mentioning is that many recent methods utilize pairwise constraints to construct a graph and apply algorithms from graph theory to solve the pose estimation problem~\cite{zhang20233d,DBLP:journals/corr/abs-1902-01534,yang2023mutual}. 
In particular, for rotation estimation, there is also a pairwise constraint~\cite{peng2022arcs,chen2022deterministic}, which can be used to decouple rotation motion. 
Therefore, it can reduce the difficulty and speed up the calculation of rotation motions~\cite{peng2022arcs}.

\subsubsection{Drawback of Decomposition Method}
Broadly speaking, the decomposition method can decompose the motion and then solve the sub-problems sequentially~\cite{yang2020teaser,peng2022arcs,liu2018efficient, Huang-TPAMI2024}. 
In particular, to solve the rotation estimation, one can solve the direction of the rotation axis and then compute the rotation angle via a geometric constraint~\cite{Huang-TPAMI2024,peng2022arcs}. 

Specifically, given two vectors, $\bm{a}$ and $\bm{b}$, 
\begin{equation}
	\mathbf{R}\bm{a}=\bm{b}\Rightarrow \bm{r}^T\left(\bm{a}-\bm{b}\right)=0
\end{equation}
where $\bm{r}\in\mathbb{S}^2$ is the rotated axis of $\mathbf{R}$. Therefore, one can formulate the objective using the right part. Consequently, the sub-problem can be formulated as a low (two) dimensional linear problem only containing the rotation axis $\bm{r}$. Consequently, one can solve $\bm{r}$ using efficient and robust linear model fitting methods~\cite{chin2022maximum,james2021linear}. Furthermore, since it is a low-dimensional problem, one can even exhaustively search the entire solution domain, that is, the unit sphere surface $\mathbb{S}^{2}$~\cite{peng2022arcs,liu2022globallyb}. Based on the obtained axis $\bm{r}$, the rotation angle will be easily estimated~\cite{peng2022arcs}. Finally, the rotation $\mathbf{R}$ can be assembled.

It seems that the 3DOF (Degrees of Freedom) rotation estimation problem is decomposed into 2D and 1D sub-problems. The dimensionality of the problem has been reduced, and the difficulty of the problem has also been reduced. However, there is no silver bullet~\cite{brooks1987no}, and there are several unavoidable drawbacks~\cite{chen2022deterministic,li2019fast} as follows,
 \begin{itemize}
 	\item \textbf{Propagation of error}. Understandably, the accurate estimation of the rotation angle relies heavily on the solution to the initial axis sub-problem. Consequently, any error in the solution to the first sub-problem will propagate backward.
 	\item  \textbf{Relaxation of objective}. For the first sub-problem, it drops the angle constraint. Therefore, the solution from sequentially solving two sub-problems is not necessarily the same as the solution to the original problem.
 \end{itemize} 
 
Theoretically, we provide a special case where this decomposition method will mostly fail. Specifically, three rotation motions are given in the same scene,
\begin{align}
	\mathbf{R}_1=\exp\left(\theta_1\left[\bm{r}_1\right]_\times\right)
	\\
	\mathbf{R}_2=\exp\left(\theta_2\left[\bm{r}_2\right]_\times\right)
	\\
	\mathbf{R}_3=\exp\left(\theta_3\left[\bm{r}_1\right]_\times\right)
\end{align}
where $\left[\cdot\right]_\times$ is to construct a skew-symmetric matrix from a vector. $\mathbf{R}_1$ and $\mathbf{R}_3$ have the same rotation axis $\bm{r}_1$, and $\mathbf{R}_2$'s rotation axis is $\bm{r}_2$. All three rotations have different rotation angles $\{\theta_1,\theta_2,\theta_3\}$. 
This case often happens in practical applications, e.g., registration of terrestrial LiDAR scan pairs~\cite{cai2019practical}. Assume there are $N_1,N_2,N_3$ observations satisfy rotation  $\mathbf{R}_1,\mathbf{R}_2,\mathbf{R}_3$, respectively. Furthermore, $N_2>N_1$ and $N_2>N_3$, which means $\mathbf{R}_2$ should have the most inliers and $\bm{r}_2$ is the axis of the to-be-estimated rotation.
However, if $N_1+N_3>N_2$, only considering the rotation axis constraint, the axis $\bm{r}_1$ will be the winner for the first axis estimation sub-problem.  which contradicts the fact that  $\bm{r}_2$ should be the axis solution of the entire rotation estimation problem.

\section{Motivation and Contributions}
Given a rotation motion, $\mathbf{R}\bm{a}=\bm{b}$, there are two facts:
\begin{enumerate}
		\item Not all rotations in $\mathbb{SO}(3)$ can rotate $\bm{a}$ to $\bm{b}$, which seems easily understandable.
		\item There is more than one rotation that satisfies the given rotation motion since there are many ways that can rotate $\bm{a}$ to $\bm{b}$.
\end{enumerate}
These two facts lead us to think about a question:
\begin{tcolorbox}
	\centering
\textbf{What areas are all eligible rotations located in?}
\end{tcolorbox}
\noindent In this paper, we try to answer this question, and based on what we obtained, we propose a robust and efficient rotation estimation algorithm.

\subsection{Contributions}
In this paper, we investigate the geometric relationship of rotation motion and make solving the rotation estimation problem more straightforward.
  
Specifically, given a rotation motion, $\mathbf{R}\bm{a}=\bm{b}$, we add some new sights from geometrical perspective:
\begin{itemize}
	\item \textbf{Quaternion circle}. The rotation satisfies the constraint should be a one-dimensional structure in $\mathbb{S}^3$. We prove that it should be a great circle of the unit quaternion sphere, and therefore we call it a quaternion circle.
	
	\item \textbf{Linearized expression}. Based on the idea of quaternion circle, the rotation motion constraint can be reformulated by a linear system of two equations without dropping any constraints and without adding any singularities. Accordingly, the rotation estimation problem can be considered as solving a system of linear equations.
	
	\item \textbf{Intersection point}. Geometrically, the rotation estimation problem can also be considered as seeking the intersection of many quaternion circles in a hypersphere surface $\mathbb{S}^3$. Furthermore, we reveal that seeking the intersection of quaternion circles in $\mathbb{S}^3$ can be reformulated as seeking the intersection of many 3-dimensional curves in $\mathbb{R}^3$.
\end{itemize}
Based on the above insights, we propose a novel rotation voting method to solve the outlier-robust rotation estimation problem. 
The contributions of this paper can be summarized as follows:
\begin{itemize}
	
	\item \textbf{Linear time complexity.} To the best of our knowledge, this is the first algorithm that can solve the famous outlier-robust rotation estimation problem with the linear time complexity $\mathcal{O}(N/\varepsilon^3)$, where $N$ is the number of inputs and $\varepsilon$ is the desired accuracy resolution. In addition, the algorithm can be easily parallelized. Notably, the proposed rotation voting method can solve the $10^6$ correspondence scale rotation estimation problem within $0.5$ seconds with the help of GPUs. 
    
	\item  \textbf{Exploring the dual geometric structure of rotation constraint.} In this paper, we explore the geometric structure of the dual space introduced by the rotation constraint, which is inspired by the famous Hough transform for detecting lines~\cite{hart2009hough}.
	Accordingly, the rotation estimation problem is transferred to finding intersection points of multiple one-dimensional curves in the rotation solution domain.
	
	\item \textbf{Extreme robustness.} Since the estimated rotation is sought in the solution domain by the exhaustive voting way, the proposed method can find the optimal solution with extreme robustness, which means the proposed method has the potential to be applied in many safety-critical applications~\cite{carlone2023estimation,bianchi2019localization}.
	
	\item \textbf{Immediate solutions for multiple rotation estimation~\cite{jin2024multi,stoffregen2019event}.} Different from many existing optimal rotation estimation methods, which can only solve a single rotation problem, our proposed rotation method can obtain multiple potential solutions simultaneously. As the to-be-solved rotations are voted and clustered in the same rotation space, it is immediately possible to classify many different rotations in the same scene, which is also known as motion segmentation~\cite{jin2024multi}.
	
\end{itemize}
\section{Method}

\subsection{Quaternion Circle}

\begin{definition}[Quaternion Circle]
	Given two unit vectors $\bm{a}$ and $\bm{b}\in\mathbb{S}^2$, they satisfy the  constraint 	 $\mathbf{R}\bm{a}=\bm{b}$, where rotation $\mathbf{R}\in\mathbb{SO}(3)$. Geometrically, The quaternion formulation of the unknown-but-to-be-solved rotation should be in a \textbf{great circle} (see the definition in~\cite{izumiya2011great,immler1936loxodrome}) in the unit quaternion sphere $\mathbb{S}^3$, and we call this one-dimensional circle as \textbf{quaternion circle} (See Fig.\ref{fig:great_circle}). 
\end{definition}
Formally, the solution of $\mathbf{R}\bm{a}=\bm{b}$ can be formulated as ~\cite{liu2022globally,liu2023absolute}
	\begin{equation}
		\mathbf{R}=\mathbf{R}_{\bm{b}}(\alpha)\mathbf{R}_{a}^{b}
	\end{equation}
	where $\mathbf{R}_{\bm{b}}(\alpha)$ is a rotation whose rotated axis is $\bm{b}$ and rotated angle is $\alpha$; $\mathbf{R}_{\bm{a}}^{\bm{b}}$ is a rotation that can rotate $\bm{a}$ to $\bm{b}$. A straight way to calculate $\mathbf{R}_{\bm{a}}^{\bm{b}}$ is rotating $\bm{a}$ to $\bm{b}$ along the minimal geodesic motion~\cite{parra2018guaranteed,liu2023absolute}. Particularly, 
	\begin{equation}
		\mathbf{R}_{{a}}^{{b}}=\exp\left(\theta\left[\bm{c}\right]_\times\right)\twoheadleftarrow
			\left[\cos\left(\frac{\theta}{2}\right),\sin\left(\frac{\theta}{2}\right)\bm{c} \right]
	\end{equation}
where $\theta\in[0,\pi]=\angle\left(\bm{a},\bm{b}\right)=\arccos\left(\bm{a}^T\bm{b}\right)$; the symbol $\twoheadleftarrow$ means the right part is the quaternion formulation of the left part;  $\bm{c}\in\mathbb{S}^2$ is the rotated axis of $\mathbf{R}_{{a}}^{{b}}$ and should satisfy $\bm{a}^T\bm{c}=0$ and $\bm{b}^T\bm{c}=0$. If $\bm{a}\neq\bm{b}$, $\bm{c}=\frac{\bm{a}\times\bm{b}}{\|\bm{a}\times\bm{b}\|}$, where $\times$ means cross product.
	 
Nonetheless, given $\bm{a}$ and $\bm{b}$,  $\mathbf{R}_{{a}}^{{b}}$ can be explicitly calculated. Therefore, there is only one unknown variable $\alpha$ for rotation $\mathbf{R}$. In other words, $\mathbf{R}$ only relies on the one-dimensional variable $\alpha$, which means given $\mathbf{R}\bm{a}=\bm{b}$, then $\mathbf{R}$ should be in a one-dimensional structure. For better understanding, we can define the one-dimensional structure as $\mathbb{C}_{quat}$. Accordingly, there is a mapping
	\begin{equation}
			\mathbf{R}(\alpha):[-\pi,\pi]\mapsto\mathbb{C}_{quat}
	\end{equation}
	by
	\begin{equation}
		\mathbf{R}(\alpha)=\mathbf{R}_{\bm{b}}(\alpha)\mathbf{R}_{a}^{b}
	\end{equation}
If there is another constraint for $\alpha$, the rotation can be solved. More specifically, if given another $\mathbf{R}\widetilde{\bm{a}}=\widetilde{\bm{b}}$, there is another one-dimensional structure for $\mathbf{R}$, then the intersection of two one-dimensional structures should be the to-be-solved rotation. It is consistent with the previous assertion~\cite{peng2022arcs,parra2018guaranteed} that two different correspondences can minimally solve the rotation estimation problem.  
\begin{tcolorbox}
	Notably, we here do not imply any two observation pairs are destined to be able to solve a rotation, because two one-dimensional structures(i.e., quaternion circles) may not necessarily intersect in $\mathbb{S}^3$~\cite{strang2012linear}, which is the incompatible case (see appendix) and rarely discussed in the literature. 
\end{tcolorbox}
	To describe the one-dimensional structure $\mathbb{C}_{quat}$ in 4-dimensional space, we introduce the unit quaternion sphere. Specifically,~\cite{ID_wiki_qua}
	\begin{align}
		\mathbf{R}_{\bm{b}}(\alpha)&\twoheadleftarrow \left[\cos\left(\frac{\alpha}{2}\right), \sin\left(\frac{\alpha}{2}\right)\bm{b}\right]
		\\
		\mathbf{R}_{\bm{a}}^{\bm{b}}&\twoheadleftarrow
		\left[q_0,\left[q_1,q_2,q_3\right]^T\right]=\left[q_0,\bm{q}_{123}\right]
	\end{align}
	where $q_0^2+q_1^2+q_2^2+q_3^2=1$.
	According to the composition of spatial rotations, which was derived by Olinde Rodrigues in 1840~\cite{rodrigues1840lois}
	\begin{align}
		\mathbf{R}(\alpha)\twoheadleftarrow\bigg[&\cos\left(\frac{\alpha}{2}\right)q_0-\sin\left(\frac{\alpha}{2}\right)\bm{b}^T\bm{q}_{123},
		\\
		&q_0\sin\left(\frac{\alpha}{2}\right)\bm{b}+\cos\left(\frac{\alpha}{2}\right)\bm{q}_{123}+\sin\left(\frac{\alpha}{2}\right)\bm{b}\times\bm{q}_{123}
		\bigg] \notag
	\end{align}
 Let $\bm{q}(\alpha)$ represent the quaternion formulation of $\mathbf{R}(\alpha)$,
\begin{align}
	\bm{q}(\alpha)=&\left[
	\begin{aligned}
		q_0
		\\
		\bm{q}_{123}
	\end{aligned}
	\right]\cos\left(\frac{\alpha}{2}\right)+\left[
	\begin{aligned}
		-\bm{b}^T\bm{q}_{123}
		\\
		q_0\bm{b}+\bm{b}\times\bm{q}_{123}
	\end{aligned}
	\right]\sin\left(\frac{\alpha}{2}\right)
	\\
	\triangleq& \bm{q}_{b1}\cos\left(\frac{\alpha}{2}\right)+\bm{q}_{b2}\sin\left(\frac{\alpha}{2}\right)
\end{align}
Observe 
\begin{align}
	&\bm{q}_{b1}^T\bm{q}_{b1}=q_0^2+q_1^2+q_2^2+q_3^2=1
	\\
	&\begin{aligned}
			\bm{q}_{b2}^T\bm{q}_{b2}=&\left(\bm{b}^T\bm{q}_{123}\right)^2+\left(q_0\bm{b}\right)^2+\left( \bm{b}\times\bm{q}_{123}\right)^2
		\\
		=&	\bm{q}_{123}^T\bm{q}_{123}\left(\cos\angle(\bm{b},\bm{q}_{123})\right)^2
		\\
		&+q_0^2+\bm{q}_{123}^T\bm{q}_{123}\left(\sin\angle(\bm{b},\bm{q}_{123})\right)^2
		\\
		=&q_0^2+q_1^2+q_2^2+q_3^2=1
	\end{aligned}
	\\
	&\bm{q}_{b1}^T\bm{q}_{b2}=-q_0\bm{b}^T\bm{q}_{123}+\bm{q}_{123}\left(q_0\bm{b}+\bm{b}\times\bm{q}_{123}\right)=0
\end{align}
%where $\angle(\bm{x},\bm{y})\in[0,\pi]$ denotes the angle between $\bm{x}$ and $\bm{y}$. 
Therefore, $\bm{q}_{b1}$ and $\bm{q}_{b2}$ are two orthonormal   bases in $\mathbb{R}^4$(see Fig.~\ref{fig:great_circle}). The structure of $\bm{q}(\alpha)$ is a one-dimensional half circle in $\mathbb{R}^4$, and its center is at $[0,0,0,0]^T$. It is worth noting that the mapping from the quaternion sphere, i.e., $\mathbb{S}^3$ to rotation space, i.e., $\mathbf{R}\in\mathbb{SO}(3)$, is a 2-to-1 mapping~\cite{hartley2009global}. In other words, $\bm{q}(\alpha)$ and $-\bm{q}(\alpha)$ represent the same rotation. Then, we have,
\begin{align}
		\bm{q}(\alpha):[-\pi,\pi]\mapsto \mathbb{C}_{quat}^+ \\ -\bm{q}(\alpha):[-\pi,\pi]\mapsto \mathbb{C}_{quat}^-\\ \mathbb{C}_{quat}^+\cup\mathbb{C}_{quat}^-=\mathbb{C}_{quat}
\end{align}
where $\mathbb{C}_{quat}^+$ and $\mathbb{C}_{quat}^-$ are symmetric half circles. Eventually, since $\|\bm{q}(\alpha)\|=1$, $\mathbb{C}_{quat}$ must be a great circle of the unit quaternion sphere.

\subsection{Closed-Form Solution For Outlier-Free Cases}\label{sec:closed form}
In practical applications, it is typical to have more than two input observations. If there are no mismatches, i.e., outliers~\cite{liu2025robustly,parra2018guaranteed},  then the rotation can be solved using least squares methods, which have been thoroughly discussed~\cite{horn1987closed,markley199930,umeyama1991least}. In this part, we provide a new geometrical perspective to solve this non-minimal outlier-free rotation estimation problem.

Geometrically, given $\bm{x}_i$ and $\bm{y}_i$, the to-be-solve rotation $\bm{q} $ should be in quaternion circle $\bm{q}_i\left(\alpha_i\right)=\bm{q}_{i,b1}\cos\left(\frac{\alpha_i}{2}\right)+\bm{q}_{i,b2}\sin\left(\frac{\alpha_i}{2}\right)$. Furthermore, if the input $\{\bm{x}_i,\bm{y}_i\}_{i=1}^N$ are all ideally \textit{clean} (i.e., no noise and outliers), all quaternion circles should intersect at a pair of antipodes (i.e., $\pm\bm{q}$) due to symmetry (see Fig.~\ref{fig:great_circle} (b)). Therefore, the rotation estimation problem can be solved by seeking the intersections of many great circles in $\mathbb{S}^3$. 

In real applications, clean inputs are really rare, and input observations typically contain unavoidable noise. To solve for the optimal solution, the least-squares objective function can be formulated as 
\begin{equation}
	\min_{\alpha_i,\bm{q}\in\mathbb{S}^3}\sum_{i=1}^{N}\|\bm{q}-\bm{q}_i\left(\alpha_i\right)\|^2\label{eqn:free-obj}
\end{equation}
Note that $\bm{q}_i\left(\alpha_i\right)$ here represents the $i$-th quaternion circle. Geometrically, $\min_{\alpha_i}\|\bm{q}-\bm{q}_i\left(\alpha_i\right)\|$ is the nearest distance between the   to-be-solve rotation $\bm{q}$ and the $i$-th quaternion circle, and the Eq.~\eqref{eqn:free-obj} is to minimize the total distance between $\bm{q}$ and  quaternion circles. However, solving Eq.~\eqref{eqn:free-obj} is to solve not only a rotation but also many $\alpha_i$, which increases difficulties. 

To eliminate $\alpha_i$, we introduce the following lemma,
\begin{lemma}
	Given $\mathbf{R}\bm{x}_i=\bm{y}_i$, $\bm{q}$ is the quaternion expression form of the to-be-solved rotation, and it is in a quaternion circle $\bm{q}_i\left(\alpha_i\right)=\bm{q}_{i,b1}\cos\left(\frac{\alpha_i}{2}\right)+\bm{q}_{i,b2}\sin\left(\frac{\alpha_i}{2}\right)$ , then there must exist 
	\begin{equation}
		\left[	\begin{aligned}
			{\bm{q}}_{i,b3}^T
			\\
			{\bm{q}}_{i,b4}^T
		\end{aligned}\right]
		\bm{q}=0
	\end{equation} 
where $\left\{\bm{q}_{i,b1},\bm{q}_{i,b2},\bm{q}_{i,b3},\bm{q}_{i,b4} \right\}$ is an orthonormal  basis of $\mathbb{R}^4$.
\label{lemma:orth basis}
\end{lemma}
The Lemma \ref{lemma:orth basis} can be easily proved, and we omit trivial explanations. . 
Geometrically speaking, Lemma~\ref{lemma:orth basis} reveals that the quaternion circle is the intersection of two hyperplanes and the unit quaternion sphere~\cite{strang2012linear}. Moreover, the normal directions of this two hyper-planes are $\bm{q}_{i,b3}$ and $\bm{q}_{i,b4}$, respectively. The hyper-planes must pass through the origin points $[0,0,0,0]^T$.

According to quaternion circle theory and Lemma~\ref{lemma:orth basis}, given $\mathbf{R}\bm{x}_i=\bm{y}_i, i=1\cdots N$, there will be $N$ quaternion circles
\begin{equation}
	\bm{q}_i\left(\alpha_i\right)=\bm{q}_{i,b1}\cos\left(\frac{\alpha_i}{2}\right)+\bm{q}_{i,b2}\sin\left(\frac{\alpha_i}{2}\right), i=1\cdots N
\end{equation}
Therefore,
\begin{equation}
	\left[
		\begin{aligned}
		{\bm{q}}_{1,b3}^T
		\\
		{\bm{q}}_{1,b4}^T
	\end{aligned}
	\right]\bm{q}=0,
		\left[
	\begin{aligned}
		{\bm{q}}_{2,b3}^T
		\\
		{\bm{q}}_{2,b4}^T
	\end{aligned}
	\right]\bm{q}=0,  \cdots ,
		\left[
	\begin{aligned}
		{\bm{q}}_{N,b3}^T
		\\
		{\bm{q}}_{N,b4}^T
	\end{aligned}
	\right]\bm{q}=0
\end{equation}
We stack the equations,
\begin{equation}
	\left[\bm{q}_{1,b3},\bm{q}_{1,b4},\cdots, \bm{q}_{N,b3},\bm{q}_{N,b4}\right]^T\bm{q}\triangleq\mathbf{Q}\bm{q}=\bm{0}
	\label{eq:stack}
\end{equation}
where $\mathbf{Q}\in \mathbb{R}^{2N\times4}$, which can be calculated from input observations. To solve $\bm{q}$ by solving Eq.~\eqref{eq:stack} in the least squares sense, it is to find the eigenvector of $\mathbf{Q^T}\mathbf{Q}$ with the smallest eigenvalue~\cite{hartley2003multiple}. In other words, we can estimate the optimal rotation by solving a system of linear equations. 

Honestly, solving  Eq.~\eqref{eq:stack}  may not be the most efficient way to calculate rotation estimation for outlier-free cases. However, it provides a new geometric perspective for calculating rotation and yields valuable analytic insights. Especially, many traditional methods that aim at solving Wahba's problem typically calculate all the observations together indiscriminately, more specifically, to calculate an attitude profile matrix $\sum_{i=1}^{N}\bm{y}_i\bm{x}_i^T$~\cite{markley2013equivalence}. By contrast, using quaternion circles can reformulate each $\mathbf{R}\bm{x}_i=\bm{y}_i$ as two linear equations without dropping any constraints and without involving any singularities, which is particularly useful for classifying inliers and outliers in outlier-contaminated cases~\cite{klivans2018efficient}.

\subsection{Rotation Voting for Outlier-Contaminated Cases }
According to the previous discussion, the rotation estimation problem can be considered as seeking the intersection problem~\cite{aiger2021efficient} between quaternion circles on the unit sphere surface in 4-dimensional space. When there are outliers, i.e., mismatches, in the inputs,  there are many intersection points, most of which correspond to \textit{fake} rotations, on the quaternion sphere surface.  Therefore, to obtain the robust rotation estimation with outlier-contaminated inputs, it can be transformed
  to seek the point that \textit{approximately intersects}~\cite{aiger2019general} the most quaternion circles.   Analogously, to robustly detect lines in a 2D image using the famous Hough transform method~\cite{hart2009hough,mukhopadhyay2015survey}, one needs to transform a point to a line in the dual space. The intersection points in the dual space correspond to potential lines in the original space.

Unfortunately, unlike line detection, there is a tricky problem for naive voting in the 4-dimensional quaternion space. Specifically, the solution domain is a hypersphere surface, which cannot be easily represented compactly. The straight way is using a 4-dimensional circumscribed hyper-cube as the accumulator space. However, it is redundant and increases the required storage space, as the degrees of freedom of rotation are inherently 3. To alleviate this problem, we explore a novel accumulator space to represent the dual space for rotation estimation. 

\subsubsection{Stereographic Projection}
 \begin{figure}
 	\centering
	\begin{tabular}{cc}
		\includegraphics[width=0.48\linewidth]{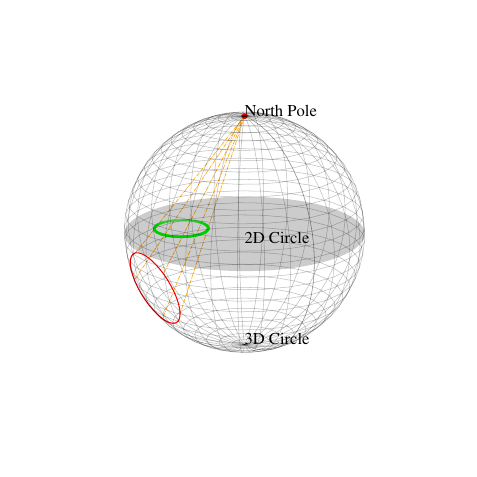}
&
			\includegraphics[width=0.48\linewidth]{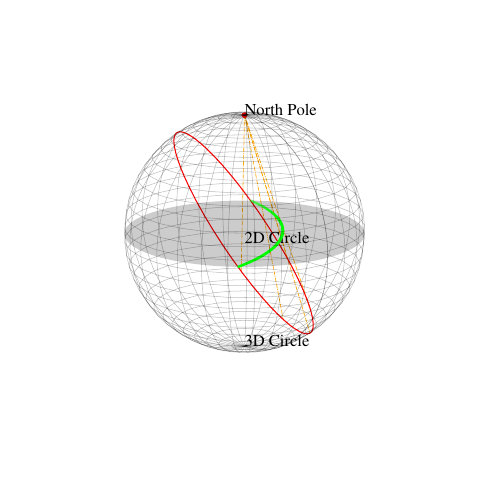}
			\\
		(a) & (b)
	\end{tabular}
	\caption{Stereographic projection in $\mathbb{R}^3$. (a) Illustration of circle preserving property. (b) Great circles in $\mathbb{S}^2$, which are analogous to quaternion circles in $\mathbb{S}^3$, are projected into circles on the equator plane $\mathbb{R}^2$. Notably, since quaternion circles are symmetrical, we only show the projected curve from the \textit{lower} half-circle.}
	\label{fig:stereo_projection}
\end{figure}

Notably, there are always two symmetric points that can \textit{approximately intersect} the most quaternion circles due to the symmetry of the unit quaternion sphere surface and quaternion circles. Therefore, taking into account the symmetry, we can reduce the solution domain to a half unit quaternion sphere surface (including the equator). In the meantime, the quaternion circles on the surface are separated into two half-circles. The remaining half circles still intersect on the half surface. Therefore, we can still find the optimal point that  \textit{approximately intersects} the most quaternion circles from the half sphere surface. At this point, although we have reduced the space of the original solution domain by half, the structure of the half-unit quaternion sphere surface is still inconvenient to handle. 

\begin{figure}[t]
	\centering
	\begin{tabular}{cc}
		\includegraphics[width=0.48\linewidth]{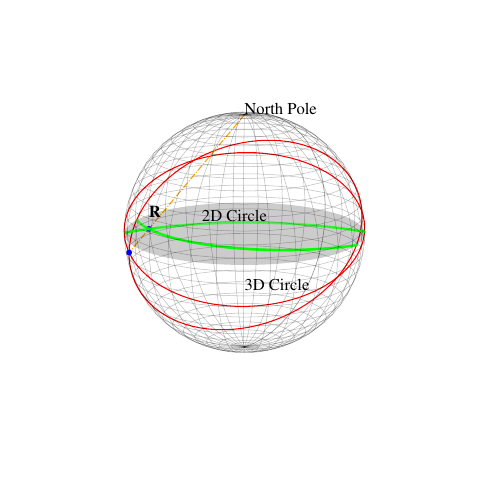}
		&
		\includegraphics[width=0.48\linewidth]{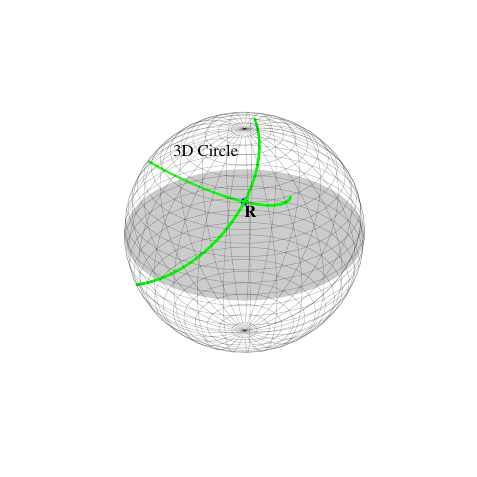}
		\\
		(a) & (b)
	\end{tabular}
	\caption{Intersections between projected circles. (a) The intersections of great circles in $\mathbb{S}^2$ are homomorphically projected into the intersections of 2D circles on the equator plane $\mathbb{R}^2$. (b) The 3D circles in $\mathbb{R}^3$ are projected from quaternion circles in $\mathbb{S}^3$. Notably, the green circles in (b) are in the ball instead of on the surface. The intersection between 3D circles corresponds to the to-to-solved rotation, which corresponds to the intersection between quaternion circles. }
	\label{fig:intersection}
\end{figure}

Furthermore, we utilize the stereographic projection~\cite{olsen2010geometry,terzakis2014quaternion}, which is able to project a half unit quaternion sphere surface (including equator) in $\mathbb{R}^4$ to a 3-dimensional solid ball, whose radius is 1, in  $\mathbb{R}^3$. As a consequence, the half quaternion circles are projected to partial circles in $\mathbb{R}^3$.  Surprisingly, these partial circles (curves) can be explicitly calculated from inputs (see Fig.~\ref{fig:stereo_projection} and~\ref{fig:intersection}). The intersection points between projected curves are also generated in the solid ball. By seeking the point that \textit{approximately intersects} the most projected curves in $\mathbb{R}^3$, we can recover the point that \textit{approximately intersects} the most quaternion circles on the quaternion sphere surface. In other words, we can find the optimal rotation in the unit solid ball in $\mathbb{R}^3$.  %Therefore, a solid box in $\mathbb{R}^3$ can cover the unit ball and be the accumulator space sufficiently. 

Formally, a stereographic projection is a perspective projection of a sphere (e.g., quaternion sphere $\mathbb{S}^3$), through a specific point on the sphere (typically, the \textit{north} pole $[0,0,0,1]^T$), onto a plane perpendicular to the diameter through the projection point (typically, the equator plane). The projection process can be 
 \begin{align}
 	\mathbf{P}(\bm{q})&=\left[\frac{q_0}{1-q_3},\frac{q_1}{1-q_3},\frac{q_2}{1-q_3}\right]^T
 	\\
 	\mathbf{P}'(\bm{p})&=\left[\frac{2p_x}{1+\bm{p}^T\bm{p}},\frac{2p_y}{1+\bm{p}^T\bm{p}},\frac{2p_z}{1+\bm{p}^T\bm{p}},\frac{\bm{p}^T\bm{p}-1}{1+\bm{p}^T\bm{p}}\right]^T
 \end{align}
where $\bm{p}=[p_x,p_y,p_z]\in\mathbb{R}^3$ is the projected point of $\bm{q}$; $\mathbf{P(\cdot)}$ is the projection process; since this projection is homeomorphic~\cite{terzakis2014quaternion}, there will be an inverse process $\mathbf{P'(\cdot)}$. 

After introducing the stereographic projection, we have 
\begin{remark}
	Given a quaternion circle $\mathbb{C}_{quat}$ in $\mathbb{S}^3$, it can be transformed to a circle or a straight line in $\mathbb{R}^3$ by stereographic projection; we denote the projected curve as $\mathbb{C}_{ste}$.  
\end{remark}
\noindent Particularly, the relation between $\mathbb{C}_{quat}$ and $\mathbb{C}_{ste}$ is following the one of the most exciting properties of stereographic projection: \textbf{circle preserving}, which claims the stereographic projection takes circles to circles and lines~\cite{olsen2010geometry} (see Fig.\ref{fig:stereo_projection}). 

Explicitly, according to Lemma~\ref{lemma:orth basis}, $\mathbb{C}_{quat}$ should satisfy 
\begin{equation}
	\bm{q}_{b3}^T\bm{q}=0, \bm{q}_{b4}^T\bm{q}=0
\end{equation} 
\begin{align}
	\bm{q}_{b3}^T=\left[{q}_{b3_0},{q}_{b3_1},{q}_{b3_2},{q}_{b3_3}\right]
\\
	\bm{q}_{b4}^T=\left[{q}_{b4_0},{q}_{b4_1},{q}_{b4_2},{q}_{b4_3}\right]
\end{align}
Using $\bm{p}$ to substitute $\bm{q}$,
\begin{align}
	\frac{1}{1+\bm{p}^T\bm{p}}\left(\left[2p_x,2p_y,2p_z,\bm{p}^T\bm{p}-1\right]\bm{q}_{b3}\right)=&0 \label{eq:quat_sphere_1}
	\\
	\frac{1}{1+\bm{p}^T\bm{p}}\left(\left[2p_x,2p_y,2p_z,\bm{p}^T\bm{p}-1\right]\bm{q}_{b4}\right)=&0\label{eq:quat_sphere_2}
\end{align}
For Eq.\eqref{eq:quat_sphere_1}, it turns out to be
\begin{align}
		\bm{p}^T\bm{p}+\left[\frac{2q_{b3_0}}{q_{b3_3}},\frac{2q_{b3_1}}{q_{b3_3}},\frac{2q_{b3_2}}{q_{b3_3}}\right]\bm{p}&=1&, {q}_{b3_3}\neq0 \label{eq:sphere_1}
		\\
		\left[q_{b3_0},q_{b3_1},q_{b3_2}\right]\bm{p}&=0&,{q}_{b3_3}=0 \label{eq:plane_1}
\end{align}
Similarly, for Eq.\eqref{eq:quat_sphere_2},
\begin{align}
		\bm{p}^T\bm{p}+\left[\frac{2q_{b4_0}}{q_{b4_3}},\frac{2q_{b4_1}}{q_{b4_3}},\frac{2q_{b4_2}}{q_{b4_3}}\right]\bm{p}&=1&, {q}_{b4_3}\neq0 \label{eq:sphere_2}
	\\
\left[q_{b4_0},q_{b4_1},q_{b4_2}\right]\bm{p}&=0&,{q}_{b4_3}=0\label{eq:plane_2}
\end{align}
Geometrically, Eq.~\eqref{eq:sphere_1} and \eqref{eq:sphere_2} represent two 3D spheres centered at $\left[\frac{-q_{b3_0}}{q_{b3_3}},\frac{-q_{b3_1}}{q_{b3_3}},\frac{-q_{b3_2}}{q_{b3_3}}\right]$ and $\left[\frac{-q_{b4_0}}{q_{b4_3}},\frac{-q_{b4_1}}{q_{b4_3}},\frac{-q_{b4_2}}{q_{b4_3}}\right]$. Their radii are $\frac{1}{|q_{b3_3}|}$ and $\frac{1}{|q_{b4_3}|}$, respectively. Similarly, Eq.~\eqref{eq:plane_1} and \eqref{eq:plane_2} represent two 3D planes whose normal directions are parallel with $\left[q_{b3_0},q_{b3_1},q_{b3_2}\right]^T$ and $\left[q_{b4_0},q_{b4_1},q_{b4_2}\right]^T$, respectively.

Consequently, if ${q}_{b3_3}=0$ and ${q}_{b4_3}=0$, $\bm{p}$ lies in a  line generated by two planes. Otherwise,  $\bm{p}$ lies in the intersected circle by two spheres or a plane and a sphere in $\mathbb{R}^3$.

\subsubsection{Rotation Voting Algorithm}
With the help of stereographic projection, seeking the rotation that intersects the most quaternion circles can be implemented as seeking a point that intersects the most spacial curves in $\mathbb{R}^3$ (see Fig.~\ref{fig:multiple-solutions}(a)), which can be voted using a 3-dim tight accumulator space, e.g., a circumscribed cube of the unit sphere in $\mathbb{R}^3$.

The rotation voting algorithm can be summarized as algorithm~\ref{alg:rotation voting}. First, we divide the unit box as a grid-like accumulator with the step size $\varepsilon$. Given observations $\{\bm{x}_i,\bm{y}_i\}_{i=1}^N$, each quaternion circle can be linearly represented by $\bm{q}_{i,{b}1}$ and $\bm{q}_{i,{b}2}$. Then, each half quaternion circle is discretized into $J$ points $\left\{ \bm{q}_i\left(\alpha_j\right)\right\}_{j=1}^J$, and every point is transferred to corresponding point $\bm{p}_i\left(\alpha_j\right)$ in $\mathbb{R}^3$ by stereographic projection.  Accordingly, the accumulator  $\mathbb{A}$ should be updated   by the location of $\bm{p}_i\left(\alpha_j\right)$.  Finally, the center of the block with the highest number in $\mathbb{A}$ will be returned as the point that intersects the most curves. Accordingly, the optimal rotation $\mathbf{R}^*$ can be recovered from the returned point.

In the rotation voting method, for each input observation, the computational complexity is stable, which is related to the sampling number $J$ and the accumulating resolution $\varepsilon$. Therefore, the entire time complexity of the rotation voting with $N$ inputs is $\mathcal{O}(N/\varepsilon^3)$. In addition, the rotation voting method can be parallelized easily, as each input can be computed individually and in parallel. Therefore, the rotation voting method can be much efficient, especially in the case of GPU acceleration for large-scale data processing.

It is worth noting that there are many modified Hough voting methods, e.g., kernel-based Hough transform~\cite{limberger2015real}, progressive probabilistic Hough transform~\cite{matas2000robust}, and deep learning-based Hough transform~\cite{zhao2021deep}, which might provide better performance, however, it is beyond the scope of this paper. In addition, since we have already linearized rotation motion by quaternion circle, there are plenty of globally optimal algorithms, which are designed for robustly solving linear systems, might be applied to solve rotation, such as~\cite{li2009consensus,chin2022maximum,liu2022globallya}.

\begin{algorithm}[t]
	 \SetKwBlock{DoParallel}{do in parallel for each $\left\{\bm{x}_i,\bm{y}_i\right\}$:}{end}
	\caption{Rotation Voting Algorithm}\label{alg:rotation voting}
	\KwIn{Observations $\{\bm{x}_i,\bm{y}_i\}_{i=1}^N$; Accumulating resolution $\varepsilon$; Sampling number $J$ for quaternion circle.}
	\KwOut{Optimal rotation $\mathbf{R}^*$.}
	Initialize an empty accumulator $\mathbb{A}$ by dividing the cube $\left[-1,1\right]^3$  with the step $\varepsilon$ in each dimension\;
	 $\bm{\alpha}=\left\{\alpha_j\right\}_{j=1}^J$ by  discretizing $\left[-\pi,\pi\right]$ uniformly\;
	\DoParallel{
		%$\mathbf{R}_{\bm{x}_i}^{\bm{y}_i}\leftarrow\left\{\bm{x}_i,\bm{y}_i\right\}$ by minimal geodesic motion\;
		Calculate basis vectors $\bm{q}_{i,{b}1}$ and $\bm{q}_{i,{b}2}$\;
		\ForEach{$\alpha_j$}
		{
			%Calculate $\bm{q}\left(\theta_j\right)$ at $\theta_j$ in quaternion cirlce \;
			$\bm{q}_i\left(\alpha_j\right)\leftarrow\bm{q}_{i,b1}\cos\left(\frac{\alpha_j}{2}\right)+\bm{q}_{i,b2}\sin\left(\frac{\alpha_j}{2}\right)$\;
			%$\bm{q}\left(\alpha_j\right) \leftarrow \mathbf{R}\left(\theta_j\right)=\mathbf{R}_{\bm{y}_i}\left(\theta_j\right)\mathbf{R}_{\bm{x}_i}^{\bm{y}_i}$\;
			$\bm{p}_i\left(\alpha_j\right)\xleftarrow{ \text{projection}}\bm{q}_i\left(\alpha_j\right)$ \;
		$\mathbb{A}\leftarrow\mathbb{A}+1$ at $\bm{p}_i\left(\alpha_j\right)$\;
		}
	}
	The center of the block with the highest number in $\mathbb{A}$ is returned, which corresponds to the to-be-sought rotation $\mathbf{R}^*$. 
\end{algorithm}

\subsection{Estimating Multiple Rotations Simultaneously}\label{sec:multiple-rotation}

Furthermore, we consider a situation where there are multiple inputs $\{\bm{x}_i,\bm{y}_i\}_{i=1}^N$, and these multiple inputs are from more than just one rotation motion $\{\mathbf{R}_i\}_{i=1}^M$, where $M$ is the number of rotations. In other words, there are multiple rotation motions in the same scene. We may need to solve multiple rotations from the inputs simultaneously~\cite{jin2024multi}. 

\begin{figure}
	\begin{tabular}{cc}
		\includegraphics[width=0.45\linewidth]{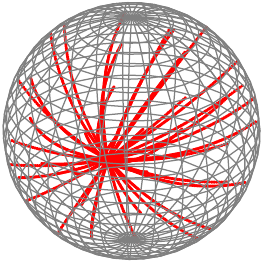}  & \includegraphics[width=0.45\linewidth]{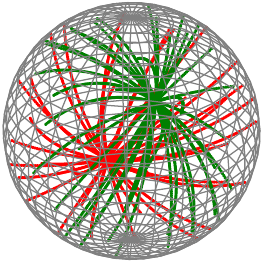} \\
		(a)&(b) 
	\end{tabular}
	\caption{Illustration of multiple rotation motions after stereographic projection. (a) There is only one rotation motion and all projected (red) curves will intersect at only one point (called the cluster point), which corresponds to the to-be-solved rotation solution. (b) If there are two rotation motions in the inputs, there will be two cluster points. All red curves belonging to one rotation movement intersect at one point. In the meantime, all blue curves intersect at another point, and they belong to another rotation.  }
	\label{fig:multiple-solutions}
\end{figure}

For example, if there are two rotation motions in the setting (see Fig.~\ref{fig:multiple-solutions}(b)), the inputs should be divided to three groups, denoted as $\mathbb{G}_1$, $\mathbb{G}_2$ and $\mathbb{G}_3$:
\begin{align}
	\{\bm{x}_i,\bm{y}_i\}_{i\in\mathbb{G}_1}&\xrightarrow{\textcircled1} \mathbf{R}_1
	\\
	\{\bm{x}_i,\bm{y}_i\}_{i\in\mathbb{G}_2}&\xrightarrow{\textcircled2} \mathbf{R}_2
		\\
	\{\bm{x}_i,\bm{y}_i\}_{i\in\mathbb{G}_3}&\xrightarrow{\textcircled3} \text{Outliers}
\end{align}

Solving this problem is far more difficult than it appears, because for each input, we need to first determine which group it belongs to. If we can correctly classify it, we can immediately calculate each rotation. However, we do not know the group information in advance. Conversely, if we could determine the rotational motion beforehand, we could instantly identify the group of each input. Thus, this problem is a well-known chicken-and-egg problem~\cite{li20073d}. Notably, the general way to solve a chicken-and-egg problem is expectation–maximization (EM) like algorithms~\cite{jin2024multi,li20073d,livsey2024applying}. However, EM-like algorithms are infamous for the risk of getting stuck in a local optimum~\cite{li20073d,Kwon2019EMCF}. In fact, sequentially solving the solutions by RANSAC is a strong choice in general cases~\footnote{\url{https://cs.nyu.edu/~fouhey/earlier/thesis/dfouhey_thesisPresentation.pdf}}. 

Surprisingly, with only minor modifications, our proposed rotation voting method can ideally solve this problem without the trouble of getting stuck in local optima.  Specifically, line 11 of algorithm~\ref{alg:rotation voting} can be modified by returning multiple possible high peaks as the solution candidates. It is well-known that generally inliers can vote cluster peaks but outliers can not~\cite {barath2019progressive,carlone2023estimation}. Therefore, the peaks in the voting space correspond to multiple possible-to-be-solved rotations. In other words, our voting-based method is naturally suited to solving the multiple-rotations estimation problem. Especially, without additional computational burden, our voting-based method solves the multiple-rotations estimation problem almost as complex as solving the unique one-rotation estimation.

\subsection{Rigid Pose Estimation}
To demonstrate the feasibility and efficiency of the proposed robust rotation voting method, we embed it into a framework to solve rigid pose estimation, which contains rotation and translation simultaneously.

Formally, the rigid pose estimation problem needs to solve rotation $\mathbf{R}\in\mathbb{SO}(3)$ and translation $\bm{t}\in \mathbb{R}^3$ from outlier-contaminated inputs. Mathematically, given $\left\{\bm{x}_i,\bm{y}_i\right\}$, the rigid motion should be 
\begin{equation}
	\mathbf{R}\bm{x}_i+\bm{t}=\bm{y}_i
\end{equation}
By introducing pairwise constraint~\cite{yang2020teaser,li2021practical}, when $i\neq j$,
\begin{align}
	\left.
	\begin{aligned}
			\mathbf{R}\bm{x}_i+\bm{t}&=\bm{y}_i
		\\
		\mathbf{R}\bm{x}_{j}+\bm{t}&=\bm{y}_{j}
	\end{aligned}
	\right\}
	\Rightarrow \mathbf{R}\left(\bm{x}_{i}-\bm{x}_{j}\right)&=\bm{y}_{i}-\bm{y}_{j}
\end{align}
Let $\bm{m}_k=\bm{x}_i-\bm{x}_{j}$ and $\bm{n}_k=\bm{y}_{i}-\bm{y}_{j}$. Then $\mathbf{R}\bm{m}_k=\bm{n}_k$, and there is only $\mathbf{R}$ should be solved. In other words, the rigid pose estimation can be transformed into a rotation estimation problem.  Consequently, we can solve robust rigid motion problems by sequentially solving rotation and translation. %Notably, some recent methods~\cite{zhang20233d,yang2023mutual,chen2023sc} utilize the same pairwise constraints to construct a graph and use approaches of graph theory to solve pose estimation problem.

It seems that we can formulate a decomposition by transforming solving a 6-dimensional problem into solving two 3-dimensional problems successively~\cite{yang2020teaser,liu2018efficient}. However, this operation has its inherent drawbacks. 
\begin{enumerate}
	\item This simple and naive operation quadratically increases the number of inputs for rotation estimation. Fortunately, this shortcoming can be mitigated by an inlier check/filter strategy described next~\cite{yang2020teaser}.
	\item The combined solution from sequentially solving sub-problems is not necessarily the same as the solution to the original problem.	Unfortunately, there is currently still no particularly good way to resolve this shortcoming~\cite{li2024transformation}.
\end{enumerate}
Nonetheless, in this paper, we focus on a robust rotation estimation algorithm, and we still use pairwise constraints to decompose rigid motion.  

It should be mentioned that our contributions to this paper are mainly in linearly solving robust rotation estimation, and we did not carefully and specially design tricks (such as multiple possible candidate solutions and graph theories~\cite{zhang20233d,yannew2023,chen2023sc}) to solve rigid pose estimation.  We just use the naive decomposition framework~\cite{li2019fast}, which perhaps is not the smartest way to solve 6D pose estimation. However, this naive decomposition way strongly relies on the rotation estimation as the fundamental part, which can aptly illustrate the superiority of linearly solving robust rotation estimation. 

\subsubsection{Inlier Check}
Given $\left\{\bm{m}_k,\bm{n}_k\right\}$, it seems we can directly apply the rotation voting algorithm to seek the optimal rotation that satisfies the most inlier observations. However, there is additional information that we can use to prune the destined outliers. Specifically, $\bm{m}_k$ and $\bm{n}_k$ typically should not be unit length, and the rotation motion will not change the length of the inputs. Therefore, 
\begin{equation}
	\lvert\|\mathbf{R}\bm{m}_k\|-\|\bm{n}_k\|\rvert=	\lvert\|\bm{m}_k\|-\|\bm{n}_k\|\rvert\left\{ 
	\begin{aligned}
		> &\mu_t,  (\text{outlier})
		\\
		\leq &\mu_t, 
	\end{aligned}
	\right.
\end{equation}
where $\mu_t$ is a preset threshold to distinguish between inliers and outliers. It conducts a rotation-invariant~\cite{yang2020teaser,li2019fast} length check. If the length before and after rotation cannot be kept consistent, it must not be an inlier, therefore, it can safely be discarded.

This strategy seems simple, however, it works especially well when the outlier rate is severely high and can remove a large number of outliers~\cite{yang2020teaser}. Notably, this inlier check strategy has been widely used in solving point cloud registration for a long time~\cite{peng2022arcs,yang2020teaser}. It is not our original contribution, and we here introduce this for completeness. 

\subsubsection{Translation Estimation}\label{sec:trans}
By conducting the inlier check and filtering many outliers, using the remaining potential correspondences, the rotation voting method can be applied to seek the optimal solution of the decoupled rotation sub-problem. After obtaining robust rotation $\mathbf{R}^*$, the following translation estimation becomes a linear model fitting problem~\cite{chin2022maximum,liu2022globallya}.
\begin{equation}
	\mathbf{R}^*\bm{x}_i+\bm{t}=\bm{y}_i
\end{equation} 
In this paper, we still use a voting-based method to implement translation estimation. Specifically, given the optimized rotation $\mathbf{R}^*$, each $\left\{\bm{x}_i,\bm{y}_i\right\}$ can calculate a candidate translation $\bm{t}_i$. After calculating all translation candidates, the optimal translation  $\bm{t}^*$ can be voted as the spatial location at which all possible solutions are most closely clustered.

\begin{figure}
\footnotesize
	\begin{tabular}{c}
	\includegraphics[width=\linewidth]{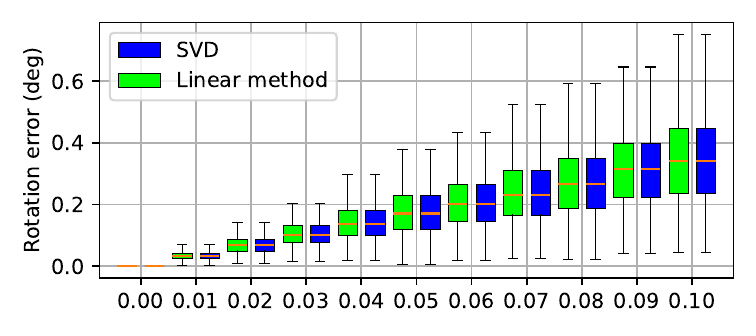}
			\\
			(a) Rotation error with different noise levels
\\
\includegraphics[width=\linewidth]{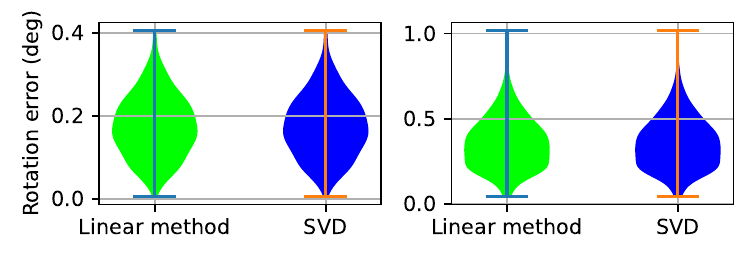}\\
\setlength{\tabcolsep}{0.24in}{\begin{tabular}{c c}
  	(b) violin plot ($\delta=0.05$) & (c) violin plot ($\delta=0.10$) 
\end{tabular}}
		\end{tabular}
	\caption{Comparison with the de facto standard outlier-free rotation estimation method, i.e., SVD~\cite{umeyama1991least}. (a) Rotation error (degree) under different noise levels, i.e., $\delta=\{0.00,\cdots,0.10\}$. (b) and (c) show the rotation error distribution under $\delta=0.05$ and $\delta=0.10$, respectively.}
	\label{fig:comparison with svd}
\end{figure}

\subsection{Multiple Rigid Pose Estimation}
Recently, some researchers have tried to solve the problem of multiple rigid pose estimation~\cite {jin2024multi}. Specifically, there are multiple to-be-solved rigid motions in the same scene.  In other words, $M$ rigid motions, $\{\mathbf{R}_i, \bm{t}_i\}_{i=1}^M$ should be solved from the input $N$ observations $\{\bm{x}_i,\bm{y}_i\}_{i=1}^N$. This case commonly occurs in dynamic scenes, in which there is normally more than one moving object and each object has its own rigid motion~\cite{rahmon2024deepftsg,jin2024multi}. For example, if there are two objects in the same scene, and each object obeys their own motion, then all the input observations can be divided to three groups,  $\mathbb{G}_1$, $\mathbb{G}_2$,  and $\mathbb{G}_3$,
\begin{align}
	\{\bm{x}_i,\bm{y}_i\}_{i\in\mathbb{G}_1}&\xrightarrow{\textcircled1} \mathbf{R}_1,\bm{t}_1
\\
\{\bm{x}_i,\bm{y}_i\}_{i\in\mathbb{G}_2}&\xrightarrow{\textcircled2} \mathbf{R}_2,\bm{t}_2
\\
\{\bm{x}_i,\bm{y}_i\}_{i\in\mathbb{G}_3}&\xrightarrow{\textcircled3} \text{Outliers}
\end{align}

Similarly, solving multiple rigid motion estimation and input classification are two mutually locked problems~\cite {jin2024multi}. If one is solved, the other one can also be solved accordingly~\cite{li20073d}. However, the solutions to them are both unknown.  In this paper, we use a decomposition approach to solve this multiple motion estimation problem. The approach is heavily based on our rotation voting method, which is just another way to show that our algorithm is superior enough that it can be used as a substrate for other high-level algorithms.

Specifically, by constructing pairwise constraints ($\bm{m}_k=\bm{x}_i-\bm{x}_{j}$ and $\bm{n}_k=\bm{y}_{i}-\bm{y}_{j}$,) and conducting inlier check,  there should be
 \begin{align}
 		\{\bm{m}_k,\bm{n}_k\}_{k\in\mathbb{G}_1}&\xrightarrow{\textcircled1} \mathbf{R}_1
 	\\
 	\{\bm{m}_k,\bm{n}_k\}_{k\in\mathbb{G}_2}&\xrightarrow{\textcircled2} \mathbf{R}_2
 	\\
 	\{\bm{m}_k,\bm{n}_k\}_{k\in\text{Other}}&\xrightarrow{\textcircled3} \text{Outliers}
 \end{align}
 Notably, only inliers can generate inliers, for example, only $\{\bm{x}_i,\bm{y}_i\}_{i,j\in\mathbb{G}_1}$, then $\{\bm{m}_k,\bm{n}_k\}_{k\in\mathbb{G}_1}$. We assume $\{\bm{m}_k,\bm{n}_k\}$ still obey geometric estimation contracts ~\cite{carlone2023estimation} and inliers will win more votes then outliers~\cite{chin2022maximum}.

After eliminating translation in rigid motion, the multiple rigid pose estimation problem turns to only estimating multiple rotations, which can be solved by the previously discussed method in Sec.~\ref{sec:multiple-rotation}. Furthermore, for the $i$-group, the corresponding translation $\bm{t}_i$ can be solved by the method described in Sec.~\ref{sec:trans}. Notably, in this paper, we focus on introducing the quaternion circle theory and a rotation voting algorithm, and we do not specially design bells and whistles. More fancy tricks, such as reducing the impact of pairwise constraints, for the specific applications, can be investigated in further work.

\section{Experiments}
To verify the feasibility of our proposed rotation estimation method, in this section, we conduct exhaustive experiments, including synthetic variable-controlled experiments and real-world dataset evaluation experiments. Some state-of-the-art robust rotation estimation methods are compared to demonstrate the superiority of the proposed algorithm.

\subsection{Experiment Setup}
All the experiments are conducted on a personal computer with an Intel i7 CPU, a GeForce RTX 4080, and 16G RAM. We implement the proposed method with Python and CuPy\footnote{\url{https://cupy.dev/}}. 
In this paper, if not specified, we set the resolution $\varepsilon=\frac{1}{180}$ and default sampling number $J=180$ for each quaternion circle.

\subsection{Verification of Closed-Form Solution}
In this part, we conduct controlled experiments using synthetic data to verify the feasibility and demonstrate the accuracy of the proposed quaternion-circle-based closed-form solution. Specifically, random rotation $\mathbf{R}_{gt}$ is generated in $\mathbb{SO}(3)$ and $N=10^3$ unit-length observations $\left\{\bm{x}_i,\bm{y}_i\right\}_{i=1}^N$ are randomly generated. To simulate noisy data, we add a Gaussian perturbation with standard deviation $\delta$ to the inputs and normalize all data again. The error of rotation  is calculated as 
\begin{equation}
	e_R=\arccos\left(\frac{1}{2}\left(\Tr(\mathbf{R}_{gt}^T\mathbf{R}_{est})-1\right)\right) 
\end{equation}
where $\mathbf{R}_{gt}$ is  ground truth; $\mathbf{R}_{est}$ is the estimated rotation; $\Tr(\cdot)$ is the trace of a square matrix.
We compare our linear closed-form method (Sec.~\ref{sec:closed form}) with the SVD-based method~\cite{umeyama1991least}. Different noise levels $\delta=\left\{0.00,\cdots,0.10\right\}$ are tested, and each experimental setting is repeated 500 times. 

The results are shown in Fig.~\ref{fig:comparison with svd}. In addition, we show the distribution of rotation error via violin plot when $\delta=\left\{0.05,0.10\right\}$. From the results, we think we can draw a brief conclusion that our quaternion-circles-based method and classical SVD-based method~\cite{umeyama1991least} are identical for solving rotation estimation problems since they are all closed-form solutions\footnote{More comprehensive conclusions should be more comprehensively tested experimentally, like~\cite{eggert1997estimating}.}.  Even if there is a difference, it can be practically negligible and could be inevitable computer roundoff errors~\cite{mortari2007optimal}. %except for inevitable computer roundoff errors.
Furthermore, our method is more than $\times5$ times slower than the SVD-based method. This is mainly due to the reformulating process of the linear system, which involves multiple eigenvector calculations. It is worth noting that there is a straightforward way to formulate the linear systems without calculating many eigenvectors, which is presented in the appendix. Using this way will run faster; however, the formulated linear systems are not necessarily independent (see the appendix). In other words, the information of the linear systems for solving the rotation problem is insufficient.

\begin{figure*}
\scriptsize
\includegraphics[width=\linewidth]{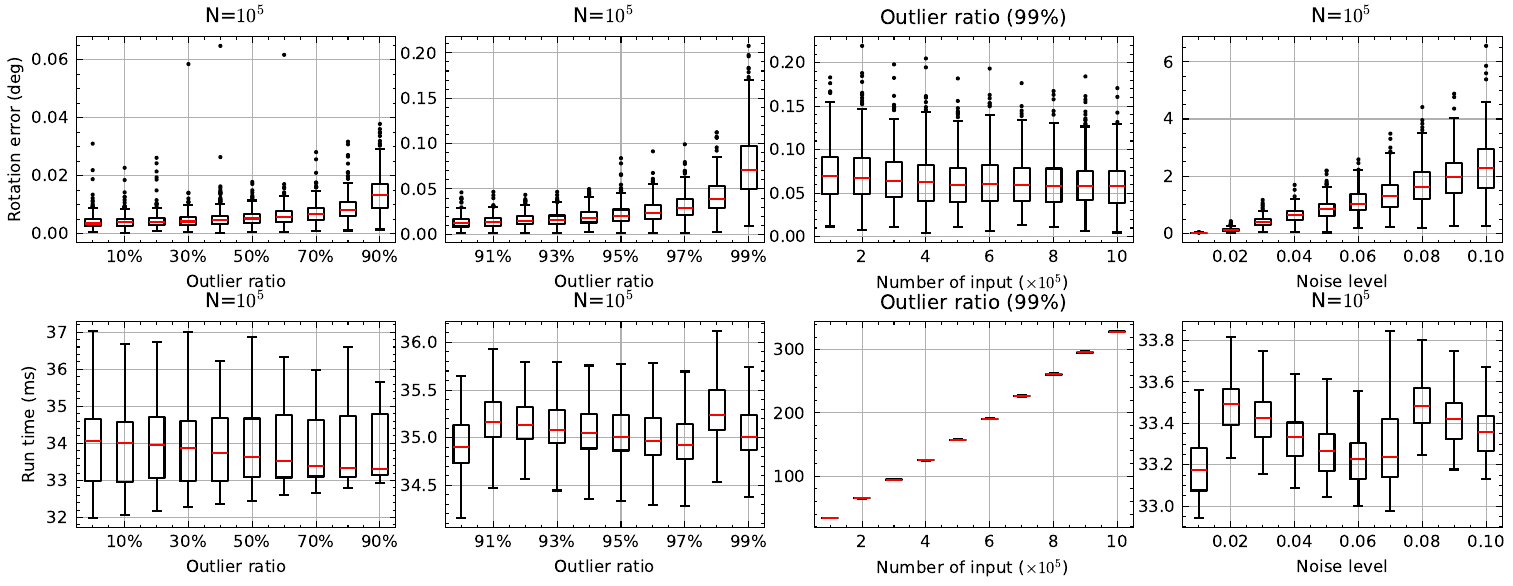}
\caption{Controlled experiments for our proposed rotation voting method. The first row shows the rotation error (degree) in different settings, and the second row shows the running time during 500 repeated experiments.}
	\label{fig:controlled experiments for RV}
\end{figure*}
\subsection{Experiments for Rotation Voting}
\subsubsection{Synthetic experiments}
To reveal the performance of our robust rotation estimation method, we conduct controlled experiments. Both running time and rotation error are recorded. Gaussian noise with standard deviation $\delta$ is added, and random mismatches are simulated as outliers. The outlier ratio is defined by $\rho=N_{outlier}/N_{total}$ where $N_{outlier}$ and $N_{total}$ are the outlier number and total input number, respectively.  To reduce randomness, each setting is repeated $500$ times. 

The settings are as follows,
\begin{itemize}
	\item \textbf{Noise $\delta$}. $N=10^5$ and $\rho=0.99$ are fixed; different noise levels $\delta=\left\{0.01,\cdots,0.10\right\}$ are tested.
	\item \textbf{Outlier ratio $\rho$}. $N=10^5$ inputs and $\delta=0.01$ is fixed; different  outlier ratios $\rho=\left\{0.10,\cdots,0.90\right\}$ are tested.
	\item \textbf{High outlier ratio $\rho$}. $N=10^5$ and $\delta=0.01$ are fixed; different outlier ratios $\rho=\left\{0.90,\cdots,0.99\right\}$ are tested.
	\item \textbf{Input number $N$}. $\delta=0.01$ and $\rho=0.99$ are fixed; different  input number $N=\left\{1 ,\cdots,10\right\}\times10^5$ are tested.
\end{itemize}
The results are shown in Fig.~\ref{fig:controlled experiments for RV}. From the results, we can draw the following conclusions,
\begin{itemize}
	
	\item As the noise level increases, the rotation error becomes larger, which is understandable. Besides, as the outlier ratio increases, the rotation error of the proposed method slightly increases.
	
	\item The proposed method can handle extreme cases with highly corrupted input data, i.e., $99\%$ outlier ratio. 
	
	\item The running time is slightly affected by the quality of input data, which demonstrates the superiority (robustness) of the proposed method.
	
	\item With enough computer memory, the running time seems linearly related to the number of inputs. This is consistent with the above statement that the complexity of the rotation voting algorithm is $\mathcal{O}(N)$.
\end{itemize}

\begin{table*}
	\centering
	\caption{Comparison with the state-of-the-art robust rotation estimation methods. $\{$Success rates $|$ average running time (second)$\}$ of various algorithms on synthetic data are listed. $10^5$ observations with $0.01$-level noise  are tested. It is considered a successful case when rotation error $e_\mathbf{R}\leq5^\circ$. $\eta={N_{os}}/{N_{total}}$, where $N_{os}$ outlier pairs are rotated around the same axis with different random angles.}
	\begin{tabular}{c|lcccccccc}
		\toprule
		\multicolumn{2}{c}{same axis ratio $\eta$} & 5\%   & 10\%  & 15\%  & 20\%  & 25\%  & 30\%  & 35\%  & 40\%  \\
		\midrule
		\multicolumn{1}{c|}{\multirow{6}[1]{*}{\rotatebox{90}{inlier ratio 20\%}}} & ARCS+O~\cite{peng2022arcs} & $100\%|1.85$ & $100\%|1.86$ & $100\%|1.88$ & $100\%|1.89$ & \colorbox{gray!15}{$99\%|1.94$} & \colorbox{gray!15}{$98\%|1.99$} & \colorbox{gray!15}{$97\%|2.02$} & \colorbox{gray!15}{$96\%|2.04$}   \\
		& ARCS+OR~\cite{peng2022arcs} & $100\%|1.89$ & $100\%|1.91$ &$100\%|1.93$ & $100\%|1.93$ & \colorbox{gray!15}{$99\%|1.99$} & \colorbox{gray!15}{$98\%|2.03$} & \colorbox{gray!15}{$97\%|2.06$} & \colorbox{gray!15}{$96\%|2.09$}  \\
		& RANSAC~\cite{fischler1981random} & $100\%|1.67$ & $100\%|1.65$ & $100\%|1.68$ & $100\%|1.66$ & $100\%|1.64$ & $100\%|1.62$ & $100\%|1.67$ & $100\%|1.66$  \\
    & GORE~\cite{parra2018guaranteed}  & \multicolumn{8}{c}{$\geq1000$ seconds} \\
		& TEASER++~\cite{yang2020teaser} & \multicolumn{8}{c}{out of memory} \\
		& OURS  & $100\%|0.04$ & $100\%|0.04$ & $100\%|0.04$ & $100\%|0.04$ & $100\%|0.04$ & $100\%|0.04$ & $100\%|0.04$ & $100\%|0.04$ \\
		\midrule
		\multicolumn{1}{c|}{\multirow{6}[2]{*}{\rotatebox{90}{inlier ratio 10\%}}} & ARCS+O~\cite{peng2022arcs} & $100\%|1.79$ & $100\%|1.82$ & \colorbox{gray!15}{$97\%|1.85$} & \colorbox{gray!15}{$96\%|1.87 $}& \colorbox{gray!15}{$97\%|1.90$} & \colorbox{gray!15}{$88\%|1.94$} & \colorbox{gray!15}{$86\%|1.98$} & \colorbox{gray!15}{$86\%|2.06$} \\
		& ARCS+OR~\cite{peng2022arcs} & $100\%|1.81$ & $100\%|1.84$ & \colorbox{gray!15}{$97\%|1.87$} & \colorbox{gray!15}{$96\%|1.90$} & \colorbox{gray!15}{$97\%|1.92$} & \colorbox{gray!15}{$89\%|1.96$} & \colorbox{gray!15}{$86\%|2.00$} & \colorbox{gray!15}{$86\%|2.08$}  \\
		& RANSAC~\cite{fischler1981random} & $100\%|1.66$ & $100\%|1.68$ & $100\%|1.68$ & $100\%|1.72$ & $100\%|1.66$ & $100\%|1.64$ & $100\%|1.64$ & $100\%|1.67$  \\
		& GORE~\cite{parra2018guaranteed}  & \multicolumn{8}{c}{$\geq1000$ seconds} \\
		& TEASER++~\cite{yang2020teaser} & \multicolumn{8}{c}{out of memory} \\
		& OURS  &  $100\%|0.04$ & $100\%|0.04$ & $100\%|0.04$ & $100\%|0.04$ & $100\%|0.04$ & $100\%|0.04$ & $100\%|0.04$ & $100\%|0.04$ \\
		\midrule
		\multicolumn{1}{c|}{\multirow{6}[2]{*}{\rotatebox{90}{inlier ratio 5\%}}} & ARCS+O~\cite{peng2022arcs} & $100\%|1.77$ & \colorbox{gray!15}{$98\%|1.79$} & \colorbox{gray!15}{$97\%|1.82$} & \colorbox{gray!15}{$90\%|1.84 $}& \colorbox{gray!15}{$85\%|1.88$} & \colorbox{gray!15}{$84\%|1.90$} & \colorbox{gray!15}{$83\%|1.95$} & \colorbox{gray!15}{$84\%|2.03$} \\
		& ARCS+OR~\cite{peng2022arcs} & $100\%|1.79$ & \colorbox{gray!15}{$98\%|1.80$} & \colorbox{gray!15}{$97\%|1.83$} & \colorbox{gray!15}{$90\%|1.85$} & \colorbox{gray!15}{$85\%|1.89$} & \colorbox{gray!15}{$84\%|1.91$} & \colorbox{gray!15}{$83\%|1.97$} & \colorbox{gray!15}{$84\%|2.05$}  \\
    		& RANSAC~\cite{fischler1981random} & \colorbox{gray!15}{$92\%|1.66$} & \colorbox{gray!15}{$94\%|1.67$} & \colorbox{gray!15}{$92\%|1.66 $}& \colorbox{gray!15}{$98\%|1.66$} & \colorbox{gray!15}{$95\%|1.66$} & \colorbox{gray!15}{$93\%|1.67$} & \colorbox{gray!15}{$92\%|1.64$} & \colorbox{gray!15}{$94\%|1.63$} \\
		& GORE~\cite{parra2018guaranteed}  & \multicolumn{8}{c}{$\geq1000$ seconds} \\
		& TEASER++~\cite{yang2020teaser} & \multicolumn{8}{c}{out of memory} \\
		& OURS  &  $100\%|0.04$ & $100\%|0.04$ & $100\%|0.04$ & $100\%|0.04$ & $100\%|0.04$ & $100\%|0.04$ & $100\%|0.04$ & $100\%|0.04$ \\
		\bottomrule
	\end{tabular}%
	\label{Tab:comparison}
\end{table*}

In addition, to highlight the robustness of the proposed method, we provide comparisons with the state-of-the-art robust rotation estimation methods, ARCS~\cite{peng2022arcs}~\footnote{\url{https://github.com/liangzu/ARCS}}, TEASER++~\cite{yang2020teaser}\footnote{\url{https://github.com/MIT-SPARK/TEASER-plusplus}}, GORE~\cite{parra2018guaranteed}~\footnote{\url{https://cs.adelaide.edu.au/~aparra/project/gore/}} and RANSAC~\cite{fischler1981random}. Concretely, we set $N=10^5$ observations with noise $\delta=0.01$  and test different inlier ratios $\{5\%,10\%,20\%\}$. Particularly, to simulate structural input outlier observations, we define the \textbf{same axis ratio} $\eta={N_{os}}/{N_{total}}$, which describes that $N_{os}$ outlier pairs are rotated around the same axis with different random angles. Each setting is repeated 200 times. Success rates and average running time are recorded. The results are shown in Table~\ref{Tab:comparison}.   

The results demonstrate the efficiency and robustness of the proposed rotation voting method. It is worth noting that when the structural observations outliers, which share the same rotation axis in the experiments, are more than inliers,  the decomposed-based methods (e.g., ARCS~\cite{peng2022arcs}) may return incorrect estimations. Although the failed situation can be somewhat rescued using other strategies (see~\cite{peng2022arcs}), however, it still drops rotation angle constraints and may lead to incorrect results.

\subsubsection{Panoramic Image Matching}
\begin{figure*}
\includegraphics[width=\linewidth]{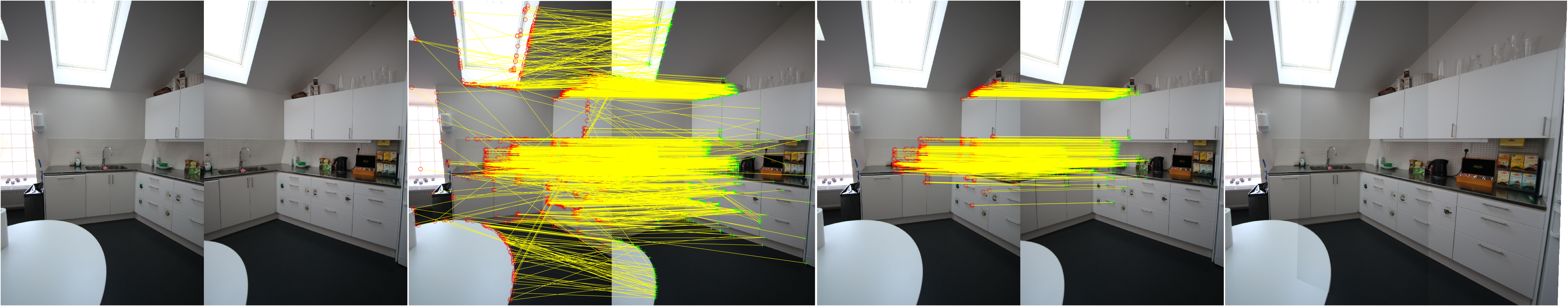}
	\begin{tabular}{>{\centering}p{4.2cm}>{\centering}p{4.2cm}>{\centering}p{4.3cm}>{\centering}p{4cm}}
		(a)&(b)&(c)&(d)
	\end{tabular}
	\caption{The working flow of rotation-only panorama stitching. (a) The input image pair. (b) Constructing the putative correspondences between image feature points and there are mismatches. (c) Removing outliers and computing the relative rotation motion. (d) Image stitching using the computed homography relationship. }
	\label{fig:panorama_stitching}
\end{figure*}
To demonstrate the practicality of the proposed rotation voting method, we test our proposed rotation voting method by conducting panoramic image matching under a rotation-only setting~\cite{yang2019quaternion,meneghetti2015image}. Specifically, we use PASSTA Datasets~\cite{meneghetti2015image}, which consists of 72 panoramic images of a \textit{Lunch Room}. 

According to the setting, between each subsequent image, the camera is rotated around the vertical axis through the optical center. Therefore, the camera has no translation motion, and the coordinates of the corresponding points across each pair of images are related by the well-known rotation-only homography~\footnote{\url{https://visionbook.mit.edu/homography.html}}. More specifically,  the relationship between the $i$-th corresponding image feature points, i.e., $\{u_i,v_i\}$ and $\{u'_i,v'_i\}$, can be 
\begin{equation}
\lambda_i    \begin{bmatrix}
	u'_i \\
	v'_i \\
	1
\end{bmatrix}
=
\mathbf{H}
\begin{bmatrix}
	u_i \\
	v_i \\
	1
\end{bmatrix}
,\quad \mathbf{H}=\mathbf{K} \mathbf{R} \mathbf{K}^{-1}
\end{equation}
where the homography matrix $\mathbf{H}$ is a 3x3 matrix but with 8 DoF (degrees of freedom) as it is estimated up to a $\lambda_i$ scale, and $\mathbf{K}$ is a 3x3  matrix of camera intrinsic parameters, which is known in advance. 

Particularly, the working flow of rotation-only panorama stitching is illustrated in Fig.~\ref{fig:panorama_stitching}. We perform pairwise image stitching every 5 images for 72 times, for example, (01.jpg to 06.jpg), (02.jpg to 07.jpg),..., (68.jpg to 01.jpg),...,(72.jpg to 05.jpg). For each pair of images, we use KAZE~\cite{chien2016use} feature\footnote{\url{https://www.mathworks.com/help/vision/ref/detectkazefeatures.html}}, which is suitable for indoor scenes, to construct the point correspondences between images. Mismatches, which might lead to seriously incorrect results, are unavoidable.  Then we perform the outlier-robust rotation estimation algorithms to obtain the homography matrix. 

\begin{figure}
	\centering
	\includegraphics[width=0.95\linewidth]{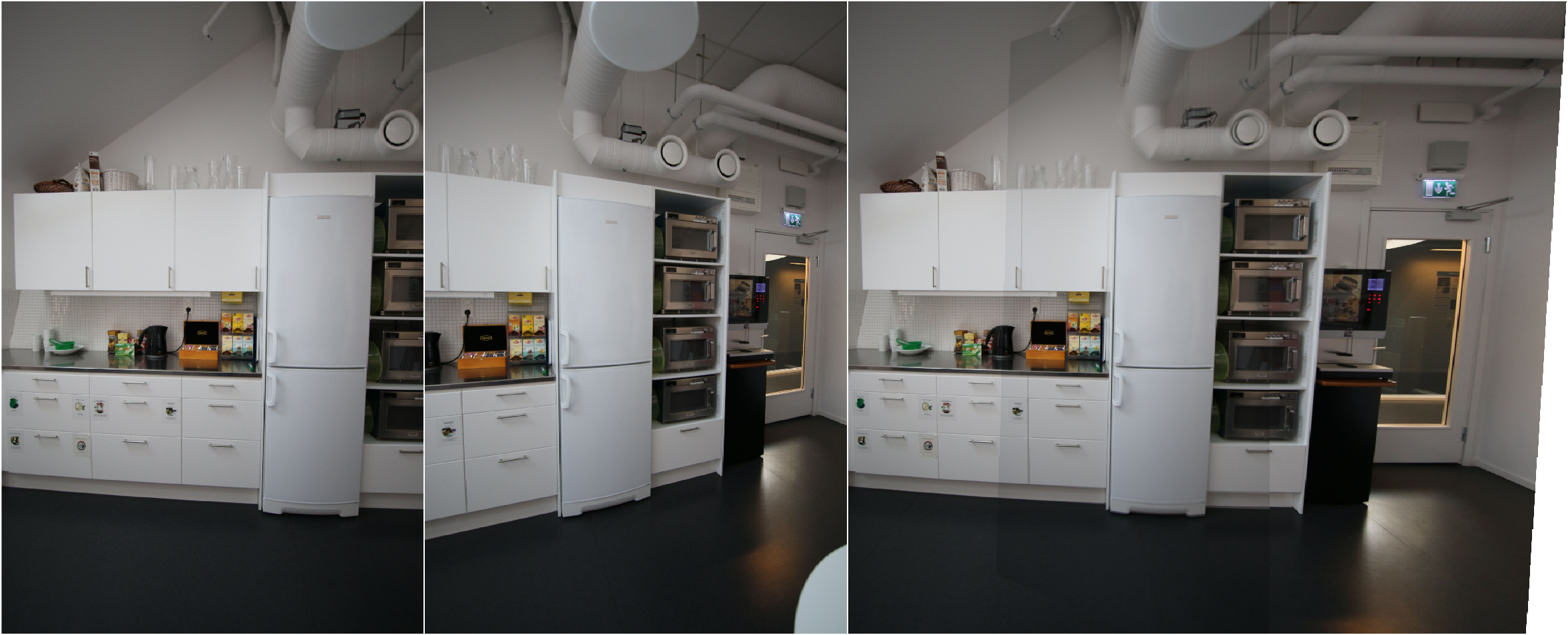}
	\\
	(a) The stitching result of images 21 and 26.
	
		\includegraphics[width=0.95\linewidth]{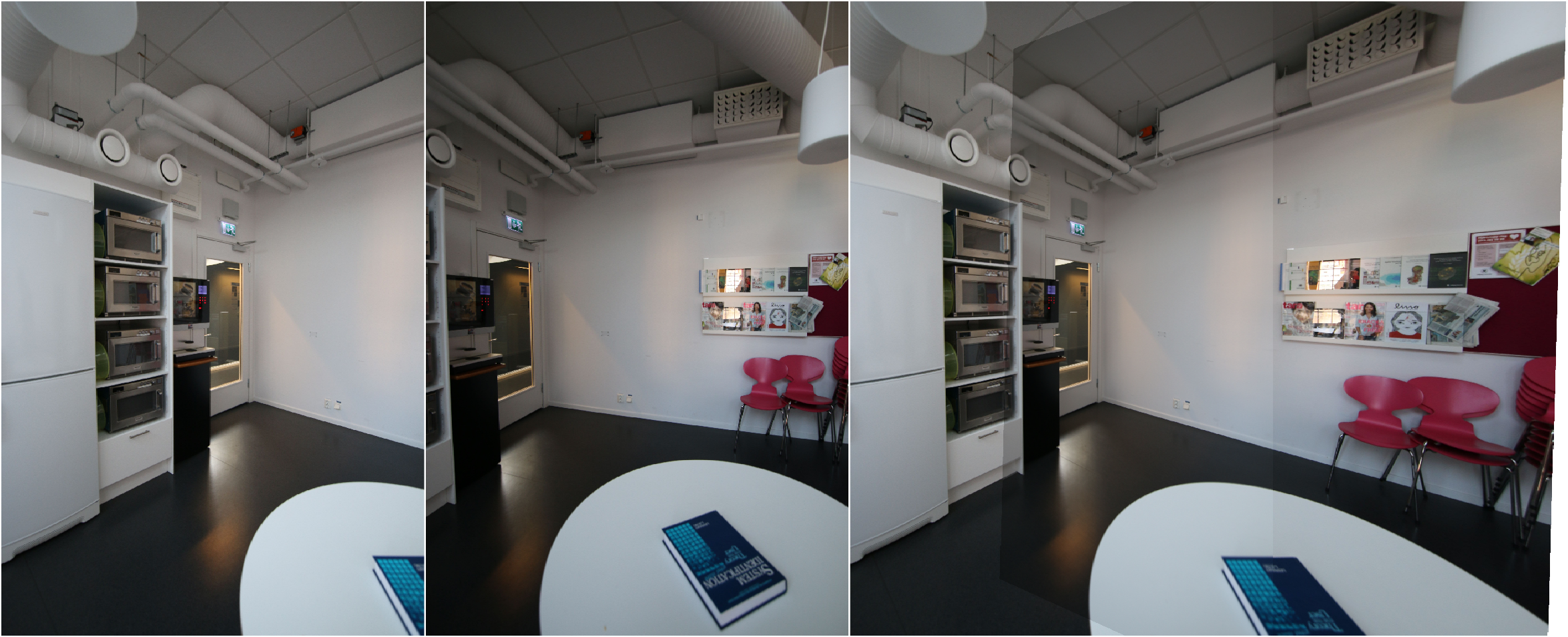}
		\\
		(b) The stitching result of images 31 and 36.
	\caption{Experimental results on passta datasets using the rotation voting method.  The left two images are inputs, and the right images are stitching results. Notably, to emphasize the results, we use a simple linear blending algorithm to combine two images. Therefore, there are obvious stitching edges in the combined images.}
	\label{fig:stitching_results}
\end{figure}

Notably, there is no ground truth in the datasets. To evaluate the results, for each pair, we manually construct a sufficient number of image key-point correspondences, and use the hand-selected point correspondences to calculate the rotation as the ground-truth. It is understandable that ground-truth is manually computed, and it should be roughly correct. Therefore, we do not report the absolute rotation error. We count the success rate, where if the absolute error is less than $3^\circ$, we think the algorithm can achieve success. Besides, we record the average running time for each algorithm. The quantitative experimental results are listed in Table~\ref {tab:stitching_results}, and visual stitching results are shown in Fig.~\ref{fig:stitching_results}.

From the results, we can find that all the methods can achieve 100\% success. In fact, the application is not very difficult for outlier-robust algorithms. Typically, each pair of images will generate about 500-2000 putative feature correspondences, in which about 50\% correspondences are inliers. All the methods can handle all the cases. However, each method requires a different running time. Among them, our rotation voting method is the most efficient because it is easily computed in parallel.

\begin{table}
	\centering
	\caption{Comparison experiments of panoramic image matching using PASSTA Datasets. Average running time and success rate are computed. }
	\begin{tabular}{lcc}
\toprule
		Methods & Average Time (Sec) & Success Rate\\
	\midrule
	ARCS+O~\cite{peng2022arcs} &0.17&100\%\\
	ARCS+OR~\cite{peng2022arcs} &0.05&100\%\\
	RANSAC~\cite{fischler1981random} &0.08&100\%\\
	GORE~\cite{parra2018guaranteed} &0.39&100\%\\
	TEASER++~\cite{yang2020teaser}&0.04&100\%\\
	OURS&\textbf{0.01}&100\%\\
	\bottomrule
		
	\end{tabular}
		\label{tab:stitching_results}
\end{table}
\subsection{Experiments for 6DOF Rigid Pose Estimation}
\begin{figure*}
	\centering
	\includegraphics[width=\linewidth]{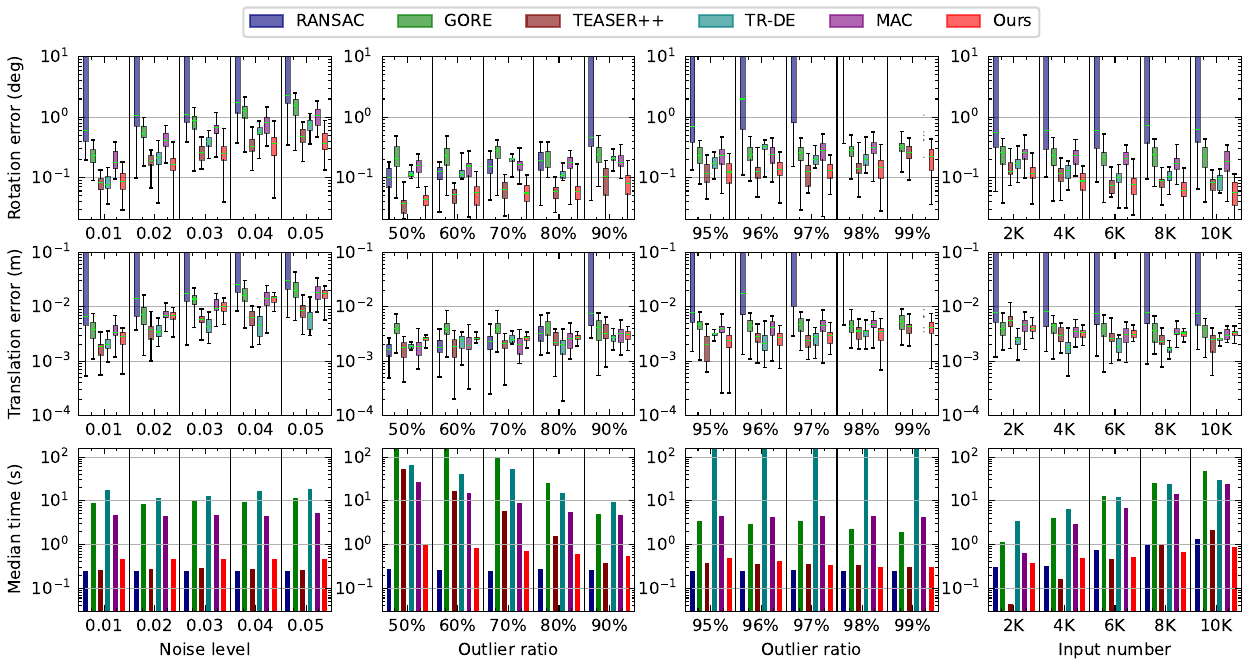}
	\caption{Comparative experiment for full rigid pose estimation using synthetic data. The top row shows the rotation error (degree) in different settings; the middle row shows the translation error (m), and the bottom row shows the running time during 500 repeated experiments. }
	\label{fig:rigid-syn}
\end{figure*}
\subsubsection{Synthetic experiments}
By using pairwise constraints, the rotation voting method can be embedded into a 6-DOF pose estimation algorithm. In this part, we first conduct controlled experiments to evaluate the performance of the proposed method.  
In addition, to evaluate the translation estimation, we define 
\begin{equation}
	e_t=\|\bm{t}_{est}-\bm{t}_{gt}\|
\end{equation}
where $\bm{t}_{gt}$ is ground-truth translation; $\bm{t}_{est}$ is estimated translation.  To simulate rigid motion, $\mathbf{R}_{gt}\in\mathbb{SO}(3)$ and $\bm{t}_{gt}\in\left[-1,1\right]$ are randomly generated. $\left\{\bm{x}_i\right\}_{i=1}^N$ are in $\left[-1,1\right]^3$, and 
\begin{equation}
\bm{y}_i=\mathbf{R}_{gt}\bm{x}_i+\bm{t}_{gt}+\delta\mathcal{N}\left(\bm{0},\mathbf{I}\right)
\end{equation}
where $\mathcal{N}\left(\bm{0},\mathbf{I}\right)$ is standard normal distribution. Outliers are generated by set $\bm{x}_i$ and $\bm{y}_i$ as random vectors in $\left[-1,1\right]^3$. The candidate domain for translation is set to $\left[-5,5\right]^3$ with discretization step size $0.025$. The experiments are repeated 500 times under the same experimental settings. 
The settings of controlled experiments are as follows,
\begin{itemize}
	\item \textbf{Noise $\delta$}. $N=5000$ and $\rho=0.90$ is fixed; different noise levels $\delta=\left\{0.01,\cdots,0.05\right\}$ are tested.
	\item \textbf{Outlier ratio $\rho$}. $N=5000$ and $\delta=0.01$ is fixed; different  outlier ratios $\rho=\left\{0.50,\cdots,0.90\right\}$ are tested.
	\item \textbf{High outlier ratio $\rho$}. $N=5000$ and $\delta=0.01$ is fixed; different  outlier ratios $\rho=\left\{0.95,\cdots,0.99\right\}$ are tested.
	\item \textbf{Input number $N$}. $\delta=0.01$ and $\rho=0.90$ is fixed; different input number $N=\left\{2,\cdots,10\right\}\times10^3$ are tested.
\end{itemize}

Moreover, we compare the proposed method with several state-of-the-art 6DOF pose estimation methods, i.e., MAC~\cite{zhang20233d}\footnote{\url{https://github.com/zhangxy0517/3D-Registration-with-Maximal-Cliques}}, TEASER++~\cite{yang2020teaser}, TR-DE~\cite{chen2022deterministic}, GORE~\cite{parra2018guaranteed}  and RANSAC~\cite{fischler1981random}. The results are shown in Fig.~\ref{fig:rigid-syn}. From the results, we can briefly draw the following conclusions,
\begin{itemize}
	\item Due to the robustness of rotation voting, the proposed method can obtain satisfactory results from highly outlier-corrupted inputs (e.g.,  $\rho=99\%$), which reveals the robustness of the rotation voting method.
	\item With the increase of noise levels $\delta$, the rotation and translation errors are all increasing, which is consistent with common sense. 

	\item As the number of inputs increases, the running time increases quadratically. This is consistent with the previous analysis: pairwise decomposition will lead to quadratic inputs for rotation voting.
	\item  Due to the inlier check procedure, as the outlier ratio increases,
	numerous outliers are removed. Therefore, the input number for rotation voting decreases accordingly, and it leads to faster running time, which is a little counter-intuitive~\cite{yang2020teaser}.
 \item Compared with the state-of-the-art 6DOF rigid pose estimation algorithms, our rotation-voting-based method, which only applies a trivial decomposition strategy, can provide comparable performance. It illustrates the high efficiency and robustness of our proposed linear rotation estimation method.
\end{itemize}

\subsubsection{Point Cloud Registration Experiments}
\begin{figure}
	\includegraphics[width=0.48\linewidth]{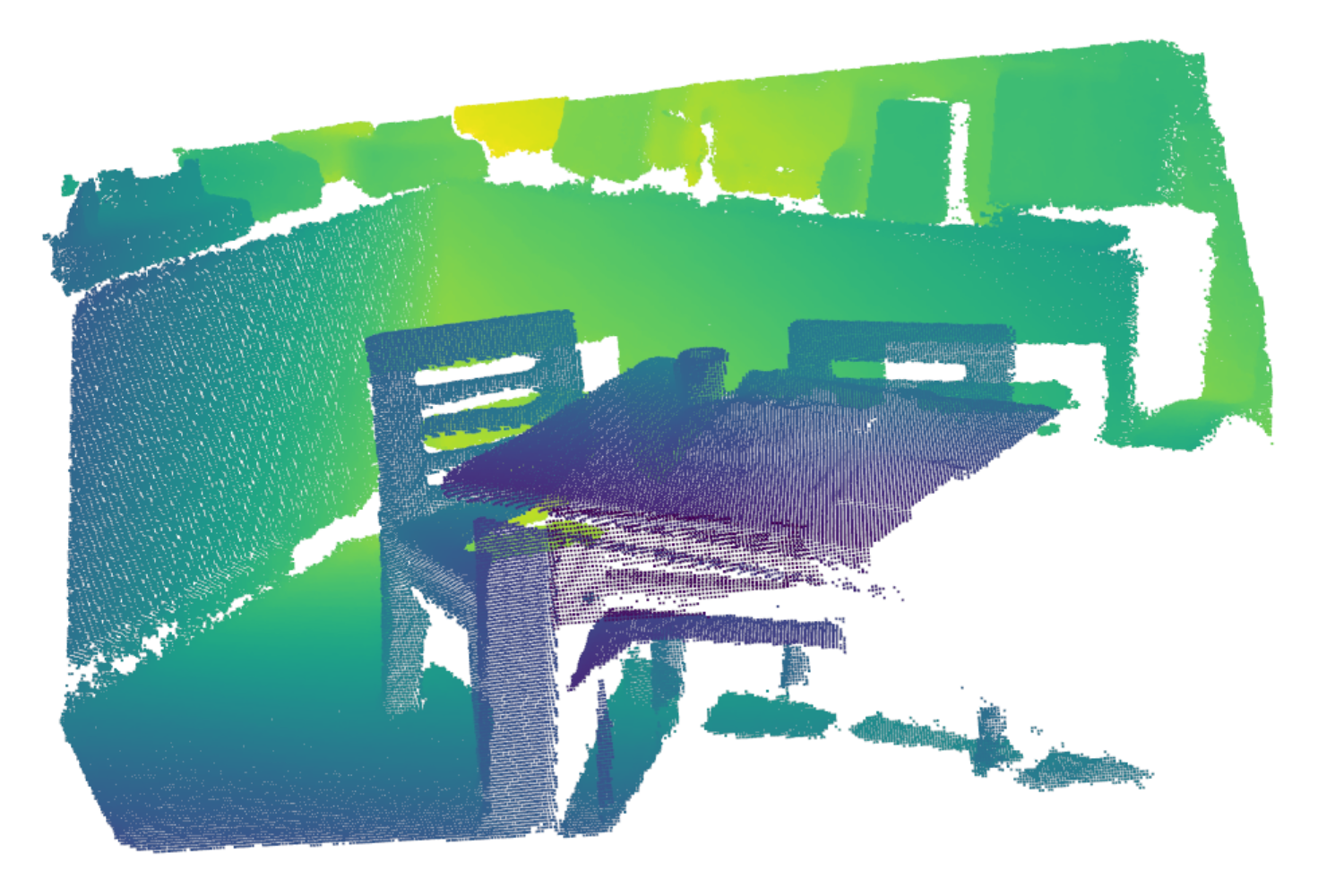}
	\includegraphics[width=0.46\linewidth]{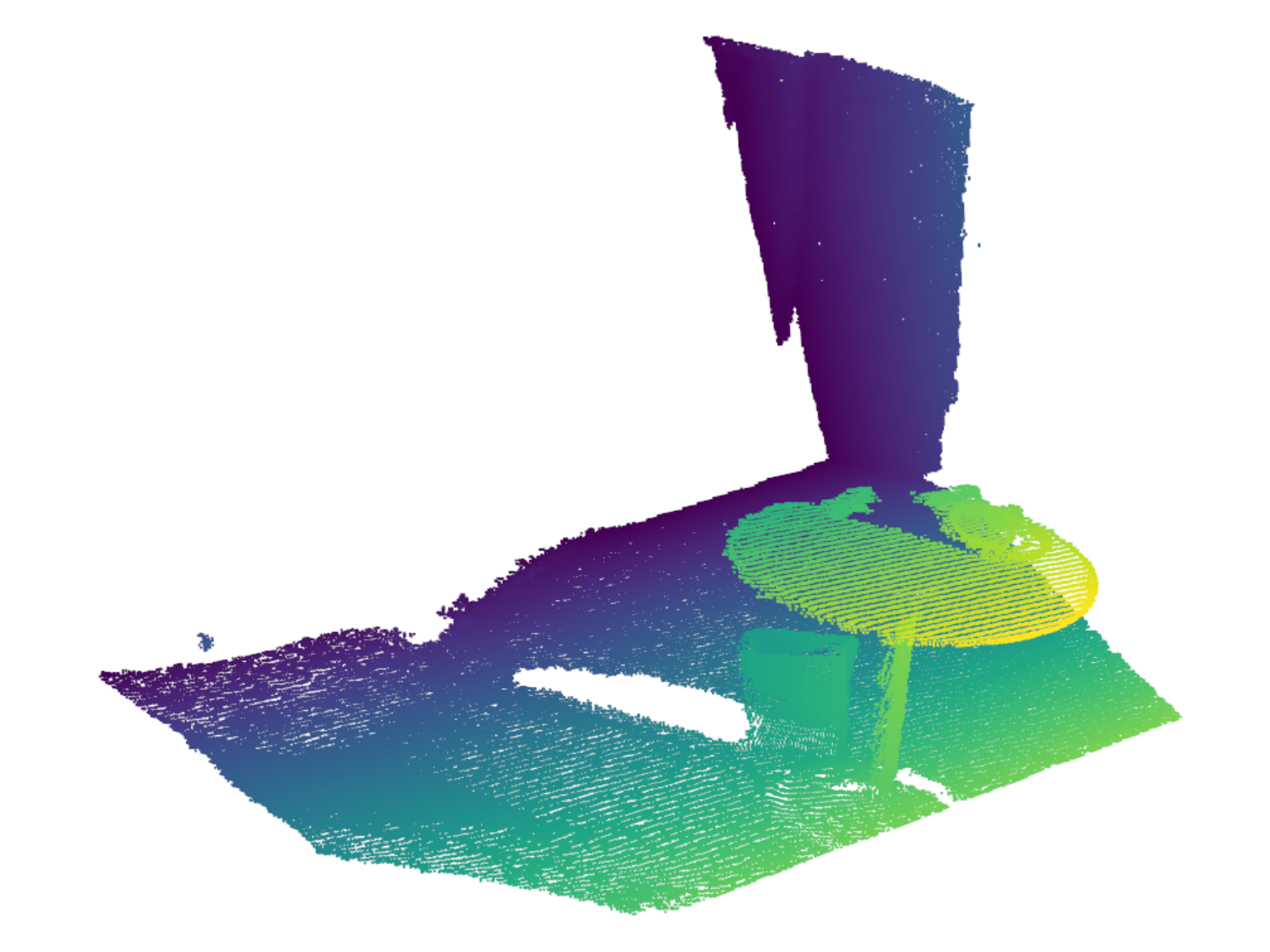}
	\\
	\begin{tabularx}{\linewidth}{>{\centering}X>{\centering}X}
	(a) 7 Scenes Red Kitchen &(b) SUN3D MIT Lab
\end{tabularx}
	\\
	\includegraphics[width=\linewidth]{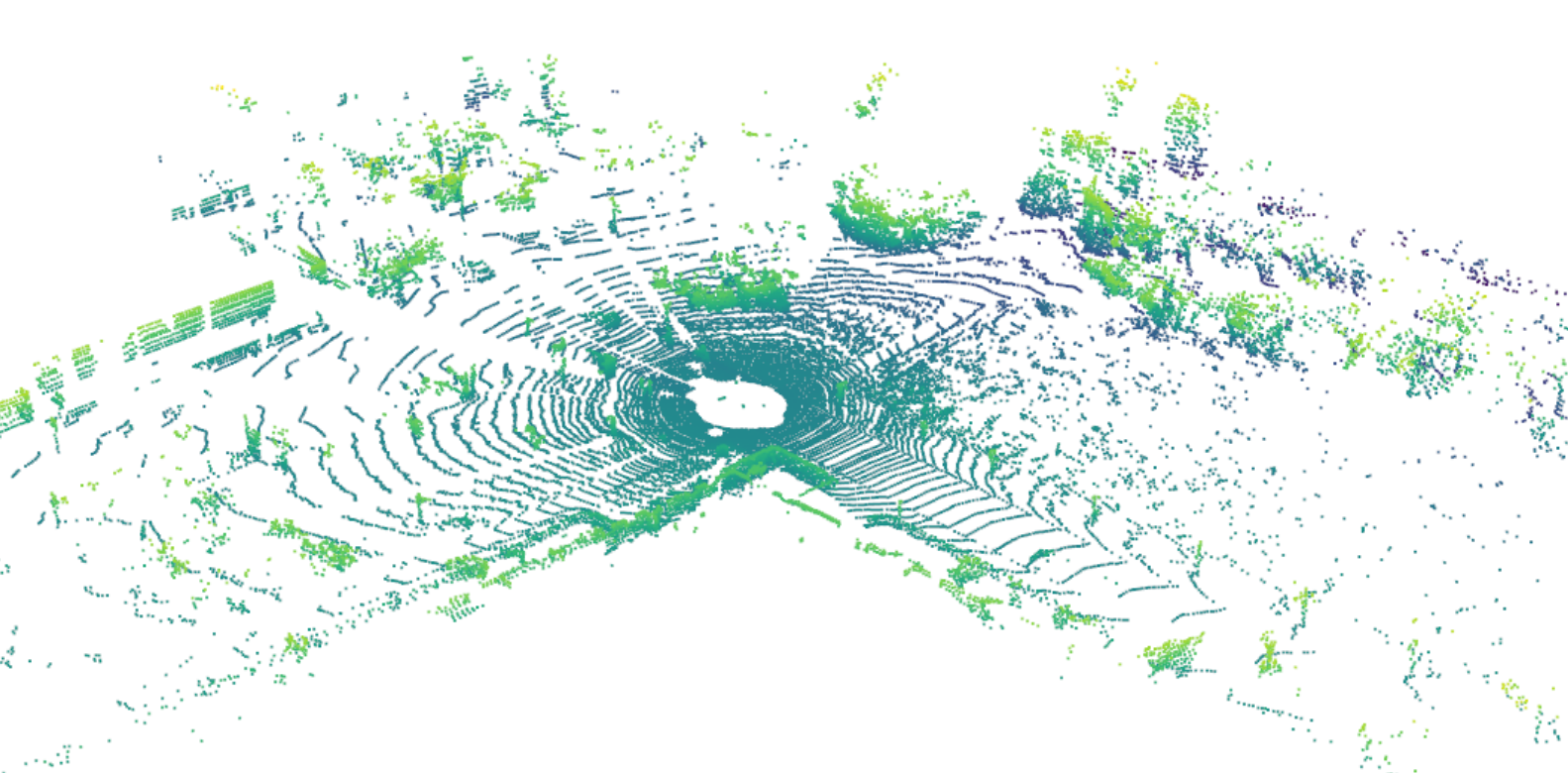}
		\begin{tabularx}{\linewidth}{>{\centering}X}
	(c) KITTI Datasets
	\end{tabularx}
	\caption{Illustrating the typical scenes for rigid point cloud registration. Scene (a) and (b) are selected from 3D(Lo)Match datasets. Most of them are indoor scenes. Point cloud (c) is randomly selected from KITTI datasets  and they are all outdoor scenes. }
	\label{fig:rigid-real}
\end{figure}
\begin{figure*}
	\includegraphics[width=\linewidth]{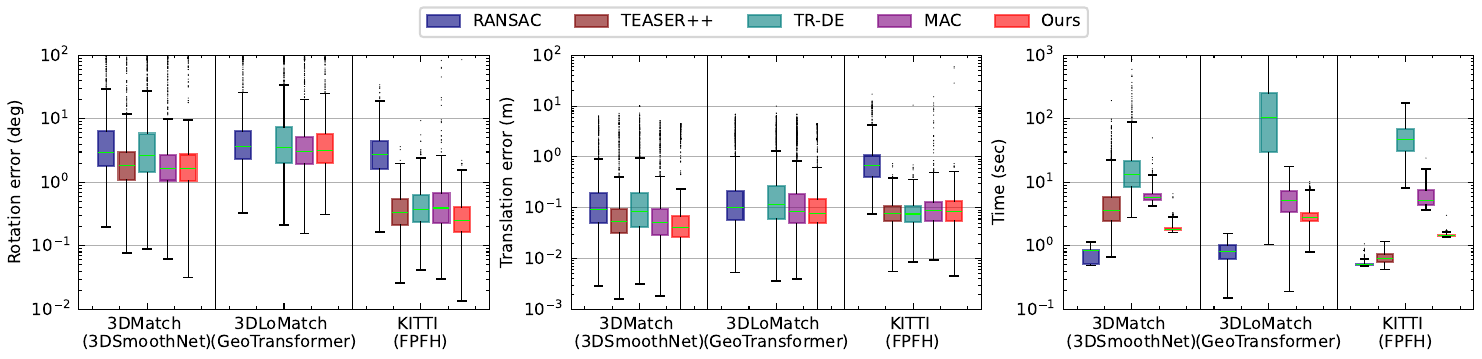}
	\caption{The 6D rigid pose estimation results on different datasets. The left figure shows the rotation accuracy. The middle figure shows the translation error. The right figure shows the duration of various methods. }
	\label{fig:point-cloud-registration}
\end{figure*}
In this part, we test the proposed method in the widely applied 3D point cloud registration application, which is to estimate rigid pose from given point clouds and is a typical pose estimation task in computer vision and robotics. Three popular real-world 3D scene datasets (see Fig.~\ref{fig:rigid-real}), i.e., 3DMatch~\cite{zeng20173dmatch}, 3DLoMatch~\cite{huang2021predator} and KITTI~\cite{geiger2012we}, are thoroughly tested. In particular, the 3DMatch  dataset mainly contains indoor scenes, and the 3DLoMatch is a low-overlap version 3DMatch dataset. In addition, the KITTI dataset is from self-driving systems and mainly contains outdoor scenes. 

We directly estimate both rotation and translation motion from pre-computed keypoint correspondences in pairs of 3D scenes. Specifically, 3DSmoothNet\cite{gojcic2019perfect}, GeoTransformer\cite{qin2022geometric} and FPFH\cite{rusu2009fast} are applied to build 3D feature correspondences for 3DMatch, 3DLoMatch and  KITTI, respectively. Both running time and pose accuracy are recorded. In addition, we also calculate the recall curves. Specifically, given a rotation threshold $\tau_{rot}$, there will be $N_{rr}$ cases where $e_{R}\leq\tau_{rot}$. Then the rotation recall is calculated by $  R_{rot}=N_{rr}/N_{total}$. Similarly, given translation threshold $\tau_{trans}$ and time threshold $\tau_{time}$, there will be $N_{rtrans}$ with $e_t\leq\tau_{trans}$ and $N_{rtime}$ with running time$\leq\tau_{time}$. The translation recall and time recall are computed by $R_{trans}=N_{rtrans}/N_{total}$ and 
$R_{time}=N_{rtime}/N_{total}$.

\begin{figure}
	\centering
	\includegraphics[width=\linewidth]{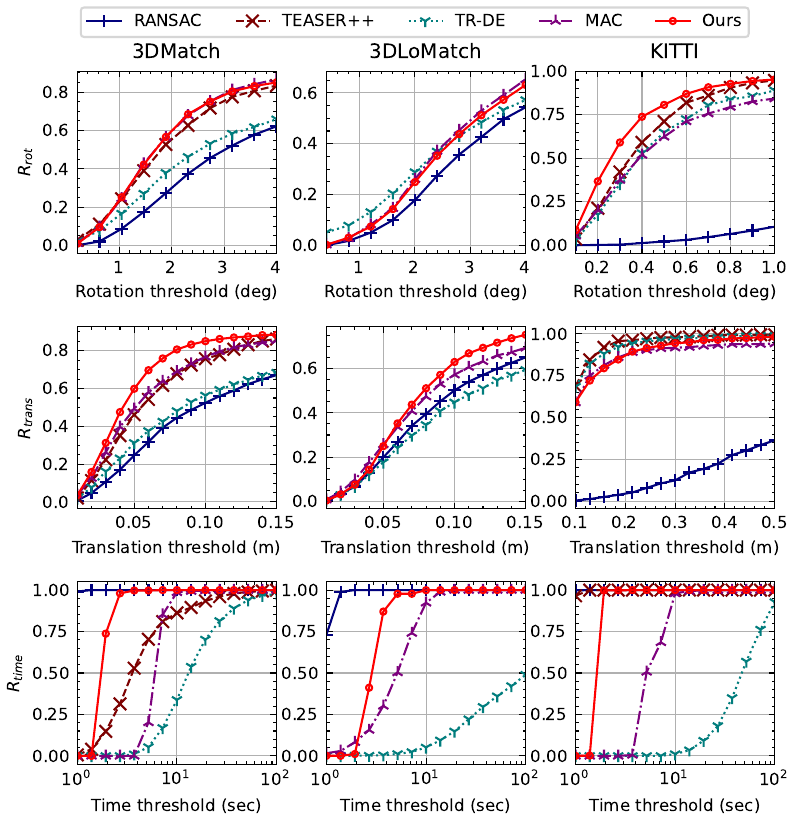}
	\caption{Recall curves on different real-world datasets (higher is better). The top row shows rotation recall for different algorithms and different datasets. The middle row shows translation recall for different algorithms and different datasets. The bottom row shows the running time recall for different algorithms and different datasets.}
	\label{fig:pcr_recall}
\end{figure}

The accuracy and time are shown in Fig.~\ref{fig:point-cloud-registration}, and the recall curves are drawn in Fig.~\ref{fig:pcr_recall}. Note that we do not show the results of the TEASER++ method on the 3DLoMatch dataset, since it cannot terminate under  1000 seconds in most cases.  Overall, although we do not pursue the best performance in 6D rigid pose estimation,  our proposed rotation voting-based method can surprisingly achieve comparable performance in terms of accuracy and efficiency compared with the state-of-the-art 6D estimation methods. Therefore, it demonstrates the power of our proposed voting-based rotation estimation method.
\subsection{Multiple Motions Estimation Experiments}
In this section, we evaluate the proposed rotation voting method to solve multiple model registration problem. In other words, there are multiple motions to be solved from the input data.

\subsubsection{Multiple Rotations Estimation}
\begin{figure}
	\centering
\includegraphics[width=\linewidth]{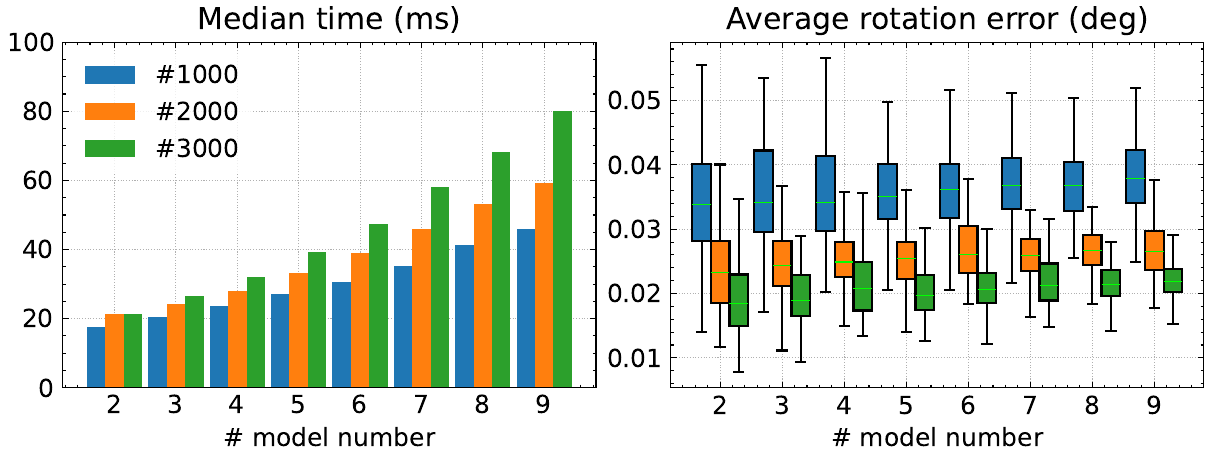}
	\caption{Performance of the rotation voting method for multiple rotation estimations with different numbers of correspondences per model, i.e., \{1000, 2000, 3000\}. Specifically, given different numbers of models, each setting is repeated 100 times. The left figure shows the median runtime, and the right figure shows the average rotation error. }
	\label{fig:multiple-rotation-results}
\end{figure}

First, we test the rotation voting method using synthetic data under only rotation motion, which demonstrate the potential of the algorithm. Specifically, random unit length vectors are randomly rotated and Gaussian noise with a standard deviation $\delta=0.01$ is added. We tested different numbers of models (rotation motions) in the same scene, i.e., \{2,...,9\}. %Frankly speaking, the problem becomes extremely difficult, even  for general multi-model estimation problems, which contain normally less than 5 models in the same scene~\cite{magri2014t,magri2019fitting,martinez2022ransac}. 
Different input correspondences, \{1000, 2000, 3000\}, per model are also tested. The rotation ground truths of each setting are randomly and uniformly distributed in the rotation domain. Particularly, the sampling number is set to $J=180\times3$ for each quaternion circle for better classification. Every experimental setting is repeated 100 times. Median running time and average rotation error are calculated, as there are multiple rotations that should be estimated, and we calculate the average error for all rotations.
The results are shown in Fig.~\ref{fig:multiple-rotation-results}.

From the results, we can draw the following conclusions,
\begin{itemize}
	\item The proposed rotation voting method can indeed solve the multiple rotation estimation problem. In addition, the method does not need careful initialization and can avoid local optimums. 
	\item As the number of models increases, the running time of the proposed method also increases.  This is because we have fixed the number of inputs in each model; thus, as the number of models increases, the total number of inputs will also increase, which will cause the total running time to increase.
\end{itemize}

Notably, the running time of these experiments is longer than previous single-model estimation ones, as the sampling number of each quaternion becomes larger for better rotation classification. 
\begin{figure*}[h]
	\centering
	\includegraphics[width=\linewidth]{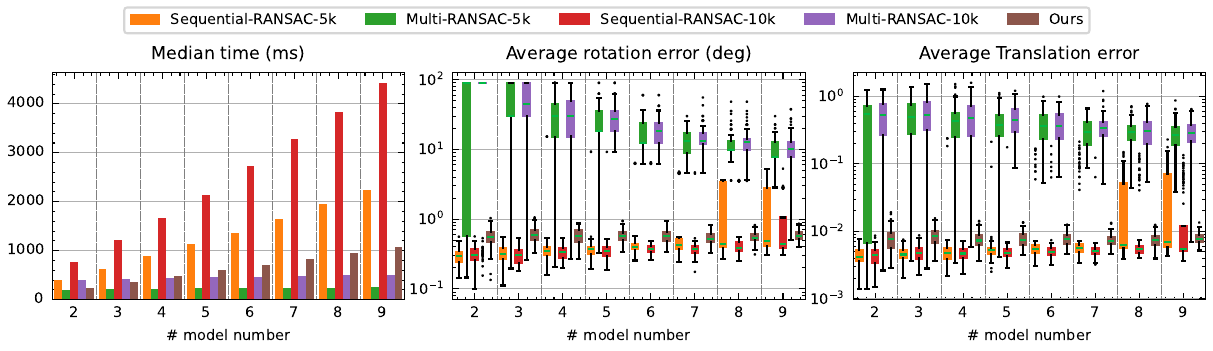}
	\caption{Multiple rigid motion estimation experiments using synthetic data. The figure on the left shows the median running time under 100 trials. The middle and right figures show the calculated errors. The model number means how many rigid motions should be estimated in the same scene.}
	\label{fig:multiple-rigid}
\end{figure*}
\subsubsection{Multiple Rigid Motions Estimation}

Furthermore, in this part, we test the proposed rotation voting method to solve the estimation of multiple rigid motions using synthetic data. To simulate multiple rigid motions, we randomly generate some models and each model has its own rigid motion. These rigid motions should be solved from the input correspondences. Specifically, each model is composed of 200 random points in the box $[-1,1]^3$. The rotation of each rigid motion is generated randomly, and the translation is also sampled randomly in $[-1,1]^3$. Gaussian noise with a standard deviation $\delta=0.01$ is added for all model data.

In addition, we compare our voting method with sequential RANSAC and multi-RANSAC, which are classical methods to solve multiple motion estimation~\cite{martinez2022ransac}. Briefly, sequential RANSAC solves one solution by the normal RANSAC algorithm and removes the inputs that belong to the solved model. The process is repeated until multiple solutions are found. Unlike that, multi-RANSAC runs once and returns several best candidates as multiple solutions. In order to fully demonstrate the performance of these RANSAC-type algorithms, we test two sampling iterations, that is, $\{5000,10000\}$. 

For each setting, the experiment is repeated 100 times. We record the running time and calculate the rotation and translation errors. The results are shown in Fig.~\ref{fig:multiple-rigid}. From the results, we can draw the following conclusions.
\begin{itemize}
	\item The proposed rotation voting method indeed can be combined with paired constraints to solve multiple rigid motion estimation problems.
	\item Comparing the RANSAC-type algorithms, the proposed voting-based method can solve the multiple solutions with significant robustness. However, RANSAC-type algorithms may return incorrect solutions, especially in a large number of models (i.e., rigid motions).  
	\item As the number of models increases, the running time also increases, which is understandable.
\end{itemize}

Why might RANSAC-type algorithms fail when the model number is large? The reason is actually simple: as the number of models increases for a specific rigid motion estimation, there are more irrelevant inputs, and the outlier rate increases. Therefore, RANSAC-type algorithms tend to fail, which is consistent with common sense. Notably, for sequential RANSAC, if the current estimation is wrong, the following results are likely to be incorrect, which is the infamous error propagation.

\subsubsection{ModelNet10 datasets experiments}
\begin{figure}
	\centering
	\includegraphics[width=\linewidth]{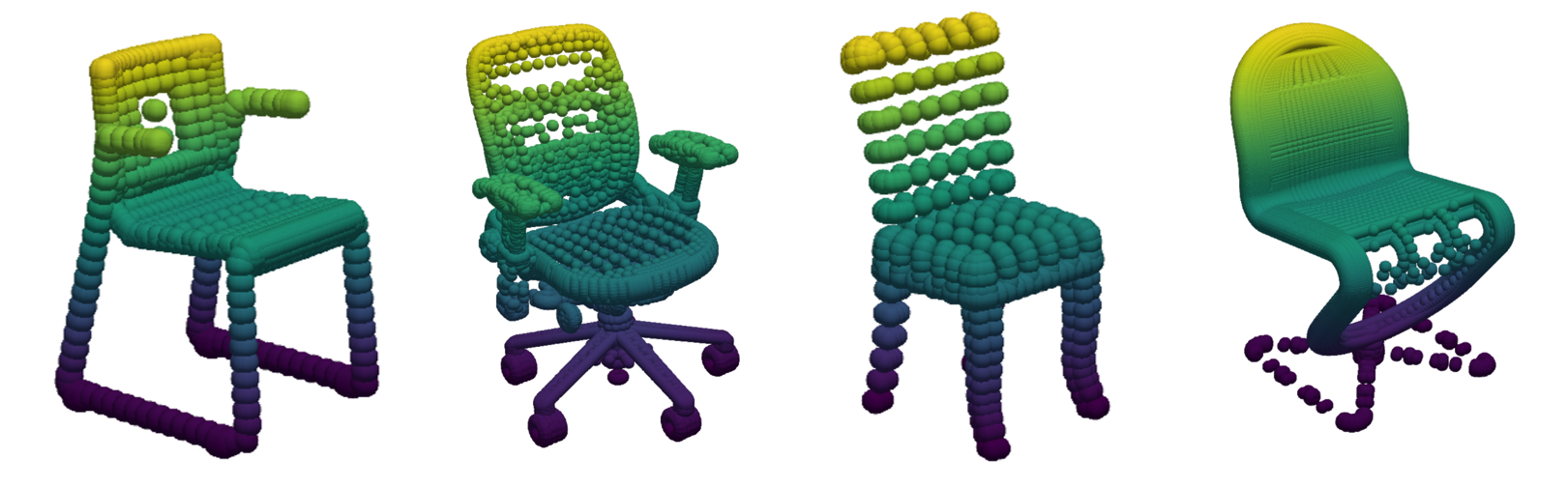}
	\caption{Typical point clouds from chair category of ModelNet10 datasets~\cite{wu20153d}. These point clouds are randomly selected to be the models in multiple rigid motion estimation. }
	\label{fig:modelnet10-sample}
\end{figure}
In this part, ModelNet10~\footnote{\url{https://modelnet.cs.princeton.edu/}} datasets (see Fig.~\ref{fig:modelnet10-sample}) are used to thoroughly evaluate the proposed method. ModelNet10 datasets contain 10 categories of point clouds generated from CAD models, and we only select the point cloud from the chair category (989 instances) to test our method. Specifically, we select a number of models from the datasets and put them in the same scene. Then we randomly rotate and translate them, and every model has its own rigid motion. Furthermore, the point correspondences are built by matching the FPFH feature description. The experimental target is to estimate all the rigid motions from the input point correspondences, which is a typical multiple-rigid motion estimation experiment.

\begin{figure}
	\centering
    \includegraphics[width=\linewidth]{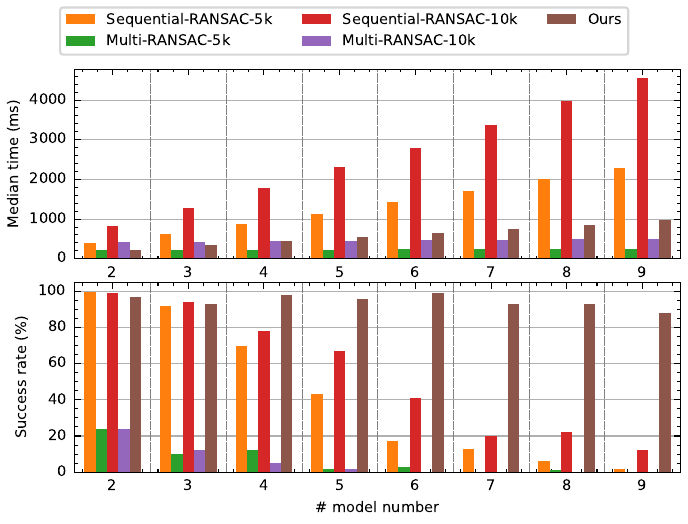}
	\caption{Multiple rigid motion estimation experiments using ModelNet10 data. The top figure shows the median time in 200 trials. The bottom figure shows the success rates for different model numbers. Only the result with average rotation error less than $2^\circ$ and average translation error than 0.02 is considered to be a success case.}
	\label{fig:modelnet10}
\end{figure}

In the experiments, we select different model numbers $\{2,\cdots,9\}$, and for each fixed number, we repeat the experiments 200 times. The average rotation and translation errors are recorded. We report the success rate and the median running time for each model number (see Fig.~\ref{fig:modelnet10}). In particular, only when $e_{\mathbf{R}}<2^\circ$ and $e_{\bm{t}}<0.02$, the case is considered correct. From the results, we can find that our voting-based method still shows high robustness. In contrast, RANSAC-type algorithms may fail, even performing well in only 2 and 3 models. It should be worth noting that even though our proposed voting-based method is quite robust,  the proposed method still cannot guarantee that the solution is correct. We should acknowledge that the problem of multiple motion estimation is more difficult than single motion estimation problem~\cite{ma2025accurate}.%, since there are ``No Free Lunch Theorems for Optimization"~\cite{wolpert1997no}, 

\section{Conclusion and Outlook}
In this paper, we propose a novel theory regarding rotation motion. Specifically, a rotation constraint can be elegantly represented by a one-dimensional quaternion circle on $\mathbb{S}^3$. Furthermore,  rotation motion can be reformulated into linear equations without dropping any constraints and without increasing any singularities.  Accordingly, the rotation estimation problem can be reformulated as seeking the intersection of many quaternion circles on $\mathbb{S}^3$,  and it can further be reformulated as seeking the intersection of many 3-dimensional partial circles in $\mathbb{R}^3$.

Inspired by Hough transformation and voting strategy in linear model fitting, we propose a rotation voting method to solve the outlier-robust rotation estimation problem. The proposed method can provide a satisfactory solution with extreme robustness. Additionally, the voting method can be easily parallelized using GPUs. The experiments demonstrate the superiority of the proposed voting algorithm in terms of efficiency and accuracy.

In summary, this paper presents an innovative and effective method for linearly addressing the robust rotation estimation problem, offering new perspectives on the problem and demonstrating significant improvements over traditional approaches. %The results of our research have important implications for enhancing rotation estimation techniques in computer vision and robotics applications. 
Furthermore, due to the existence of the linear formula for rotation motion, we strongly believe there will be a (similar or not) linear formulation for full 6D rigid motion, whose solution domain is $\mathbb{SE}(3)$. Suppose it can be discovered with more novel insights. In that case, the outlier-robust solution for 6D pose estimation may be solved using an efficient linear way similar to the one in this paper.

\ifCLASSOPTIONcaptionsoff
  \newpage
\fi

\bibliographystyle{IEEEtran}
\bibliography{IEEEabrv,./bib/paper}

\clearpage
\appendices
\onecolumn

\section{Alternative way to obtain quaternion circle}

In this part, we provide an alternative way to derive the quaternion circle, and it is inspired from~\cite{shuster1981three,horn1987closed,jia2008quaternions} and~\cite{goldman2011understanding}. 
		
		Given $\bm{a}=\left[a_1,a_2,a_3\right]^T, \bm{b}=\left[b_1,b_2,b_3\right]^T\in\mathbb{S}^2$ and they satisfy
		\begin{equation}
			\mathbf{R}\bm{a}=\bm{b}
		\end{equation}
and~\cite{kuipers1999quaternions}
\begin{equation}
	\mathbf{R}=
	\begin{bmatrix}
		q_0^2+q_1^2-q_2^2-q_3^2 & 2\left(q_1q_2-q_0q_3\right) & 2\left(q_1q_3+q_0q_2\right)\\
		2\left(q_1q_2+q_0q_3\right) & q_0^2-q_1^2+q_2^2-q_3^2  & 2\left(q_2q_3-q_0q_1\right)\\
		2\left(q_1q_3-q_0q_2\right) & 2\left(q_2q_3+q_0q_1\right) & q_0^2-q_1^2-q_2^2+q_3^2  
	\end{bmatrix}
\end{equation}
where $\bm{q}=\left[q_0,q_1,q_2,q_3\right]$ is the quaternion expression of $\mathbf{R}$ , and $\bm{q}^T\bm{q}=1$.

Since $\bm{a}$ and $\bm{b}$ are on the unit sphere surface, then
\begin{equation}
	\mathbf{R}\bm{a}=\bm{b}\Leftrightarrow \angle\left(\mathbf{R}\bm{a}, \bm{b}\right) =0 \Leftrightarrow \bm{b}^T\mathbf{R}\bm{a}=1
\end{equation} 
Substitute $\mathbf{R}$,
\begin{equation}
\bm{b}^T	\begin{bmatrix}
	q_0^2+q_1^2-q_2^2-q_3^2 & 2\left(q_1q_2-q_0q_3\right) & 2\left(q_1q_3+q_0q_2\right)\\
	2\left(q_1q_2+q_0q_3\right) & q_0^2-q_1^2+q_2^2-q_3^2  & 2\left(q_2q_3-q_0q_1\right)\\
	2\left(q_1q_3-q_0q_2\right) & 2\left(q_2q_3+q_0q_1\right) & q_0^2-q_1^2-q_2^2+q_3^2  
\end{bmatrix}\bm{a}	=1
\end{equation}
We then reformulate the equation~\cite{jia2008quaternions}, 
\begin{equation}
	\bm{q}^T
	\begin{bmatrix*}[r]
	0& -a_1&-a_2&-a_3	\\
	a_1&0&a_3&-a_2\\
	a_2&-a_3&0&a_1\\
	a_3&a_2&-a_1&0\\
	\end{bmatrix*}^T
		\begin{bmatrix*}[r]
	0& -b_1&-b_2&-b_3	\\
b_1&0&-b_3&b_2\\
b_2&b_3&0&-b_1\\
b_3&-b_2&b_1&0\\
	\end{bmatrix*}
	\bm{q}\triangleq\bm{q}^T
	\mathbf{M}
	\bm{q}=1 
\end{equation}
To calculate the eigenvalue of symmetric matrix $\mathbf{M}$,
\begin{equation}
	\det\left(\mathbf{M}-\lambda\mathbf{I}\right)=0 \Rightarrow\left(\lambda^2-1\right)^2=0
\end{equation}
where $\mathbf{I}\in\mathbb{R}^{4\times4}$ is an identity matrix. 

Consequently, the eigenvalue of $\mathbf{M}$ is $\Lambda=\diag\left(\left[1,1,-1,-1\right]\right)$. We denote the eigenvector of $\mathbf{M}$ is $\mathbf{O}\in\mathbb{R}^{4\times4}$ and $\mathbf{O}$  is an orthogonal matrix $\mathbf{O}\mathbf{O}^T=\mathbf{I}$. Accordingly,
\begin{equation}
	\mathbf{M}=\mathbf{O}
	\begin{bmatrix*}[r]
		1&0&0&0\\
		0&1&0&0\\
		0&0&-1&0\\
		0&0&0&-1\\
	\end{bmatrix*}
	\mathbf{O}^T=
	\mathbf{O}\Lambda\mathbf{O}^T
\end{equation}
Observe, 
\begin{equation}
	\bm{q}^T\mathbf{M}\bm{q}=1\Leftrightarrow \bm{q}^T\mathbf{O}\Lambda\mathbf{O}^T\bm{q}=1
\end{equation}
Let $\bm{\Gamma}=\left[\Gamma_0,\Gamma_1,\Gamma_2,\Gamma_3\right]^T=\mathbf{O}^T\bm{q}$. Then 
\begin{align}
	\left.
	\begin{aligned}
		\bm{\Gamma}^T\bm{\Gamma}= \bm{q}^T\mathbf{O}\mathbf{O}^T\bm{q}=1
		\\
	\bm{q}^T\mathbf{O}\Lambda\mathbf{O}^T\bm{q}=\bm{\Gamma}^T\Lambda\bm{\Gamma}=1
	\end{aligned}
	\right\}
	\Rightarrow
	\left.
	\begin{aligned}
		\Gamma_0^2+\Gamma_1^2+\Gamma_2^2+\Gamma_3^2=1
		\\
		\Gamma_0^2+\Gamma_1^2-\Gamma_2^2-\Gamma_3^2=1
	\end{aligned}\right\}
	\Rightarrow
	\bm{\Gamma}=\left[\Gamma_0,\Gamma_1,0,0\right]^T, \text{and} \quad\Gamma_0^2+\Gamma_1^2=1
\end{align}
Therefore, 
\begin{equation}
	\bm{q}=\mathbf{O}\begin{bmatrix}
		\Gamma_0 \\
		\Gamma_1\\
		0\\
			0
	\end{bmatrix},  \quad \Gamma_0^2+\Gamma_1^2=1
\end{equation}
Geometrically, it means $\bm{q}$ lies in a quaternion circle in $\mathbb{R}^4$.

On the other hand,
\begin{equation}
\mathbf{O}^T	\bm{q}=\begin{bmatrix}
		\Gamma_0 \\
		\Gamma_1\\
		0\\
		0
	\end{bmatrix},  \quad \Gamma_0^2+\Gamma_1^2=1
\end{equation}

The bottom two rows clearly show that every $\mathbf{R}\bm{a}=\bm{b}$ can generate two independent linear equations regarding $\bm{q}$.

In summary~\cite{liu2025robustly}, given $\bm{a}$ and $\bm{b}$, one can calculate $\mathbf{M}$, and its eigenvalue turns out to be  $\left[1,1,-1,-1\right]$. The satisfied rotation $\bm{q}$ lies in the quaternion circle, which is spanned by the two orthogonal eigenvectors corresponding to eigenvalue $1$. Moreover, $\bm{q}$ is orthogonal to the other eigenvectors corresponding to eigenvalue $-1$.

\subsection{Special Example}
Furthermore, we here directly provide  an eigenvector matrix of $\mathbf{M}$, which can be verified by mathematical software, e.g., MATLAB\footnote{\url{https://www.mathworks.com/}} and SymPy\footnote{\url{https://www.sympy.org/en/index.html}}. The matrix is however not orthogonal, but can be easily orthogonalized by Gram–Schmidt process~\cite{cheney2009linear}.
\begin{equation}
	\text{eigenvalue:}
		\begin{bmatrix*}[r]
		1&0&0&0\\
		0&1&0&0\\
		0&0&-1&0\\
		0&0&0&-1\\
	\end{bmatrix*},
	\text{eigenvector:}
\begin{bmatrix}
		-(a_3+ b_3)&  a_2+ b_2& -(a_3- b_3)&  a_2- b_2\\
		-(a_2- b_2)& -(a_3- b_3)& -(a_2+b_2)& -(a_3+b_3)\\
		a_1-b_1 &0&a_1+b_1&0\\
		0&a_1- b_1&0&a_1+b_1
	\end{bmatrix}	\label{eq:quat-original}
\end{equation} 

Let us define $\bm{e}_1,\bm{e}_2,\bm{e}_3,\bm{e}_4$, 
\begin{equation}
	\bm{e}_1=\begin{bmatrix}
		-(a_3+ b_3)\\
		-(a_2- b_2)\\
			a_1-b_1\\
			0
	\end{bmatrix},
	\bm{e}_2=
	\begin{bmatrix}
		 a_2+ b_2\\
		 -(a_3- b_3)\\
		 0\\
		 a_1- b_1
	\end{bmatrix},
	\bm{e}_3=
	\begin{bmatrix}
		-(a_3- b_3)\\
		-(a_2+b_2)\\
		a_1+b_1\\
		0
	\end{bmatrix},
	\bm{e}_4=
	\begin{bmatrix}
		a_2- b_2\\
		-(a_3+b_3)\\
		0\\
		a_1+b_1
	\end{bmatrix}
\end{equation}
When $\bm{e}_1$ and $\bm{e}_2$ are independent, $\|\bm{e}_1\|\neq0$ and $\|\bm{e}_2\|\neq0$, we can apply Gram–Schmidt process to $\bm{e}_1$ and $\bm{e}_2$, 
\begin{equation}
	\bm{\beta}_1=\dfrac{\bm{e}_1}{\|\bm{e}_1\|},\quad 
	\widetilde{\bm{\beta}}=\bm{e}_2-\bm{e}_2^T\bm{\beta}_1\bm{\beta}_1,\quad 
	\bm{\beta}_2=\dfrac{\widetilde{\bm{\beta}}}{\|\widetilde{\bm{\beta}}\|}
\end{equation}
%(2) When $a_1= b_1$, $\bm{e}_1$ is parallel to $\bm{e}_2$, and $\bm{\beta}_1^T\bm{\beta}_2=0$ cannot be constructed. Fortunately, in this case,
%\begin{equation}
%	\bm{\beta}_1=[1,0,0,0]^T,
%	\quad
%	\bm{\beta}_1=[0,b_1,b_2,b_3]^T
%\end{equation}
\faArrowCircleRight The quaternion circle can be
\begin{tcolorbox}
 \begin{equation}
 	\bm{q}=\cos(\chi)\bm{\beta}_1+\sin(\chi)\bm{\beta}_2, \quad \chi\in\left[-\pi,\pi\right]
 \end{equation}
\end{tcolorbox}
\noindent When $\bm{e}_3$ and $\bm{e}_4$ are independent,  $\|\bm{e}_3\|\neq0$ and $\|\bm{e}_4\|\neq0$,
\begin{equation}
	\bm{\beta}_3=\dfrac{\bm{e}_3}{\|\bm{e}_3\|},\quad 
	\widehat{\bm{\beta}}=\bm{e}_4-\bm{e}_4^T\bm{\beta}_3\bm{\beta}_3,\quad 
	\bm{\beta}_4=\dfrac{\widehat{\bm{\beta}}}{\|\widehat{\bm{\beta}}\|}
	\label{eq:span}
\end{equation}
%(2) When $a_1+b_1= 0$, $\bm{e}_3$ is parallel to $\bm{e}_4$, and $\bm{\beta}_3^T\bm{\beta}_4=0$ cannot be constructed. Fortunately, in this case,
%\begin{equation}
%	\bm{\beta}_3=[1,0,0,0]^T,
%	\quad
%	\bm{\beta}_4=[0,b_1,b_2,b_3]^T
%\end{equation}
\faArrowCircleRight  The linear system can be 
\begin{tcolorbox}
\begin{equation}
	\begin{bmatrix}
		\bm{\beta}_3&\bm{\beta}_4
	\end{bmatrix}^T
	\bm{q}=\bm{0}\label{eq:orth}
\end{equation}
\end{tcolorbox}
\noindent Note that arbitrary $\left\{\bm{e}_1,\bm{e}_2,\bm{e}_3,\bm{e}_4\right\}$ is not necessarily independent.%, and $\left\{\bm{\beta}_1,\bm{\beta}_2,\bm{\beta}_3,\bm{\beta}_4\right\}$ is not necessarily an orthogonal matrix.

\section{The relationship with  rotation axis constraint}
Recently, a rotation constraint is introduced and applied in solving outlier-robust pose estimation problem by decomposition way~\cite{peng2022arcs,chen2022deterministic}. Specifically~\cite{gong2011optimal},
\begin{equation}
	\mathbf{R}\bm{a}=\bm{b}\Rightarrow\bm{r}^T\left(\bm{a}-\bm{b}\right)=0 
	\label{eq:axis constraint-0}
\end{equation}
where $\bm{r}$ is rotation axis of $\mathbf{R}$.
Furthermore,
\begin{equation}
		\bm{r}^T\begin{bmatrix}
		a_1-b_1&a_2-b_2&a_3-b_3
	\end{bmatrix}^T
=0\label{eq:axis constraint}
\end{equation}
Observe
\begin{equation}
			\begin{bmatrix}
		\cos\left(\dfrac{\theta}{2}\right)& \sin\left(\dfrac{\theta}{2}\right)\bm{r}^T
	\end{bmatrix}
		\begin{bmatrix}
			0\\
		a_1-b_1\\a_2-b_2\\a_3-b_3
	\end{bmatrix}=0
	\triangleq
	\bm{q}^T		
	\begin{bmatrix}
		0\\
		\bm{a}-\bm{b}
	\end{bmatrix}=0
\end{equation}
where $\theta$ is the rotation angle of $\mathbf{R}$. It only includes one linear equation for $\bm{q}$ and drops the other one.

In addition~\cite{mortari2000second},
\begin{equation}
	\mathbf{M}	
	\begin{bmatrix}
		0\\
		\bm{a}-\bm{b}
	\end{bmatrix}=-1\cdot	
		\begin{bmatrix}
		0\\
		\bm{a}-\bm{b}
		\end{bmatrix}
\end{equation}
Therefore, 	$\begin{bmatrix}0 &
	\bm{a}-\bm{b}\end{bmatrix}^T$ is an eigenvector of $\mathbf{M}$, who is corresponding to eigenvalue $-1$.

Notably, when $\bm{a}-\bm{b}=\bm{0}$, then $\begin{bmatrix}0 &
	\bm{a}-\bm{b}\end{bmatrix}^T$  is a trivial solution, and this case is called singularity in~\cite{mortari2000second}.  In other words, when $\bm{a}-\bm{b}\rightsquigarrow\bm{0}$, Eq.~\eqref{eq:axis constraint} provides no information about $\bm{r}$, while it still stands. Exceptionally, in this case, $\bm{r}^T\bm{a}=\bm{r}^T\bm{b}=1$. 
	
	In summary, using this way to exactly solve the rotation estimation problem not only drops the angle constraint but also adds the singularity. 
%	Therefore,
%	\begin{equation}
%		\mathbf{R}\bm{a}=\bm{b} \Rightarrow\left\{ 
%		\begin{aligned}
%			\bm{r}^T\left(\bm{a}-\bm{b}\right)=0
%			\\
%			\bm{r}^T\bm{a}=\bm{r}^T\bm{b}=1
%		\end{aligned}
%		\right.
%	\end{equation}

\section{The relationship with optimal linear attitude estimator~\cite{mortari2007optimal}}
We notice that there is already a linear transformation for rotation motion~\cite{mortari2007optimal,wu2022generalized,barfoot2023certifiably}.  Specifically, it applies Cayley transformation~\cite{sarabandi2019cayley} to represent rotation matrix. 

Rodrigues vector (sometimes called Gibbs vector) is defined~\cite{hartley2013rotation},
\begin{equation}
	\bm{g}=\dfrac{1}{q_0}
	\begin{bmatrix}
		q_1
		\\
		q_2
		\\
		q_3
	\end{bmatrix}
	=
	\dfrac{1}{\cos\left(\dfrac{\theta}{2}\right)}\bm{r}\sin\left(\dfrac{\theta}{2}\right)=\bm{r}\tan\left(\dfrac{\theta}{2} \right)
\end{equation}
Clearly, when $\theta\rightsquigarrow\pi$, $\bm{g}\rightsquigarrow\infty$, it will has the singularity. Nonetheless, according to Cayley transformation~\cite{frank2017modern,DrMadibabaiasl},
\begin{align}
	\mathbf{R}&=\left(\mathbf{I}-\left[\bm{g}\right]_\times\right)^{-1}\left(\mathbf{I}+\left[\bm{g}\right]_\times\right)
	\\
	\left[\bm{g}\right]_\times&=\left(\mathbf{R}+\mathbf{I}\right)^{-1}\left(\mathbf{R}-\mathbf{I}\right)
\end{align}
where $\left[\cdot\right]_\times$ is to construct a skew-symmetric matrix from a vector. Cayley transformation is closely related to gnomonic projection~\cite{hartley2013rotation,Cayley_transform}, which projects all the points on the quanternion sphere surface from the center of the sphere, (0, 0, 0, 0), onto a tangent (3-dimensional) hyper-plane. 

\begin{equation}
	\mathbf{R}\bm{a}=\bm{b}\Rightarrow \left(\mathbf{I}-\left[\bm{g}\right]_\times\right)^{-1}\left(\mathbf{I}+\left[\bm{g}\right]_\times\right)\bm{a}=\bm{b} 
\Rightarrow \left[\bm{g}\right]_\times\left(\bm{a}+\bm{b}\right)=-\left(\bm{a}-\bm{b}\right)
\end{equation}
Equivalently,
\begin{equation}
	\left[\bm{a}+\bm{b}\right]_\times\bm{g}=\bm{a}-\bm{b}\label{eq:Cayley}
\end{equation}
There are only two linear independent equations from above  equation. Therefore, given $N$ observations, a system of equations can be formulated by simply stacking. 

Geometrically, Eq.~\eqref{eq:Cayley} provides more information about rotation than Eq.\eqref{eq:axis constraint}, and Eq.\eqref{eq:axis constraint} can be easily obtained from  Eq.~\eqref{eq:Cayley}.
\begin{equation}
	\bm{g}^T\left[\bm{a}+\bm{b}\right]_\times\bm{g}=0\Rightarrow\bm{g}^T\left(\bm{a}-\bm{b}\right)=0 \Rightarrow \bm{r}^T\left(\bm{a}-\bm{b}\right)=0
\end{equation}

When $\bm{a}+\bm{b}=0$, Eq.~\eqref{eq:Cayley} cannot provide effective constraint for $\bm{g}$: even though we know $\chi=\pi$, no information about rotation axis can be obtained, which is the singularity of  Eq.~\eqref{eq:Cayley} formulation.

In terms of formality, Eq.\eqref{eq:Cayley} is different from our proposed homogeneous linear equations. However, it can be transformed  into a homogeneous equation,
\begin{equation}
	\left[\bm{a}+\bm{b}\right]_\times\bm{g}+\bm{b}-\bm{a}=\bm{0}\Rightarrow\begin{bmatrix}
			\left[\bm{a}+\bm{b}\right]_\times&\bm{b}-\bm{a}
	\end{bmatrix}\begin{bmatrix}
	\bm{g}\\
	1
	\end{bmatrix}=\bm{0}
\end{equation}
Therefore,
\begin{equation}
	\begin{bmatrix}
		\bm{b}-\bm{a}&\left[\bm{a}+\bm{b}\right]_\times
	\end{bmatrix}\begin{bmatrix}
		1\\ \bm{g}
	\end{bmatrix}=\bm{0}\Rightarrow	\begin{bmatrix}
	\bm{b}-\bm{a}&\left[\bm{a}+\bm{b}\right]_\times
	\end{bmatrix}\begin{bmatrix}
	q_0\\
	q_1\\
	q_2\\
	q_3
	\end{bmatrix}=\bm{0}
\end{equation}
Consequently,
\begin{equation}
\begin{bmatrix}
	-\left(a_1-b_1\right)&0 &-\left(a_3+b_3\right)&a_2+b_2
	\\
	-\left(a_2-b_2\right) &a_3+b_3&0&-\left(a_1+b_1\right)
	\\
-\left(a_3-b_3\right)&-\left(a_2+b_2\right)&a_1+b_1&0
\end{bmatrix}
		\begin{bmatrix}
	q_0\\
	q_1\\
	q_2\\
	q_3
	\end{bmatrix}=\bm{0}\label{eq:quat-trans}
\end{equation}
Finally, Eq.~\eqref{eq:quat-trans} is transformed into our quaternion circle based equations, see Eq.~\eqref{eq:quat-original}.

\section{Some Useful Equations}
By collecting the above equations, we have 
\begin{equation}
	\begin{bmatrix}
		0&a_1-b_1&a_2-b_2&a_3-b_3
		\\
		-\left(a_1-b_1\right)&0 &-\left(a_3+b_3\right)&a_2+b_2
		\\
		-\left(a_2-b_2\right) &a_3+b_3&0&-\left(a_1+b_1\right)
		\\
		-\left(a_3-b_3\right)&-\left(a_2+b_2\right)&a_1+b_1&0
	\end{bmatrix}
	\begin{bmatrix}
		q_0\\
		q_1\\
		q_2\\
		q_3
	\end{bmatrix}=\bm{0}\Leftrightarrow	\begin{bmatrix}
0& \left(\bm{a}-\bm{b}\right)^T
\\
-\left(\bm{a}-\bm{b}\right)&\left[\bm{a}+\bm{b}\right]_\times
	\end{bmatrix}
	\begin{bmatrix}
	q_0\\
	q_1\\
	q_2\\
	q_3
	\end{bmatrix}=\bm{0}
\end{equation}
This formulation has been explored in dual quaternion formality with pairwise constraint~\cite{srivatsan2016estimating,arun2019registration}. 

Accordingly, we can select two independent rows and orthonormalize them to formulate a system of linear equations
\begin{equation}
\begin{bmatrix}
	\bm{\beta}_{orth_1}&	\bm{\beta}_{orth_2}
\end{bmatrix}^T	\bm{q}=\bm{0}
\end{equation}
Notably, due to singularity and linear dependence, arbitrary two rows may not be directly used in real applications.

Moreover, all four rows are eigenvectors of $\mathbf{M}$ and they are all corresponding to eigenvalue $-1$.
\begin{equation}
	\mathbf{M}\begin{bmatrix}
		0& -\left(\bm{a}-\bm{b}\right)^T
		\\
		-\left(\bm{a}-\bm{b}\right)&\left[\bm{a}+\bm{b}\right]_\times^T
	\end{bmatrix}=\begin{bmatrix}
		0& -\left(\bm{a}-\bm{b}\right)^T
		\\
		-\left(\bm{a}-\bm{b}\right)&\left[\bm{a}+\bm{b}\right]_\times^T
	\end{bmatrix}\begin{bmatrix}
		-1&0&0&0\\
		0&-1&0&0\\
		0&0&-1&0\\
		0&0&0&-1
	\end{bmatrix}
\end{equation}
In addition, there is a similar formulation for eigenvalue 1.
\begin{equation}
	\mathbf{M}\begin{bmatrix}
		0& -\left(\bm{a}+\bm{b}\right)^T
		\\
		-\left(\bm{a}+\bm{b}\right)&\left[\bm{a}-\bm{b}\right]_\times^T
	\end{bmatrix}=\begin{bmatrix}
		0& -\left(\bm{a}+\bm{b}\right)^T
		\\
		-\left(\bm{a}+\bm{b}\right)&\left[\bm{a}-\bm{b}\right]_\times^T
	\end{bmatrix}\begin{bmatrix}
		1&0&0&0\\
		0&1&0&0\\
		0&0&1&0\\
		0&0&0&1
	\end{bmatrix}
\end{equation}
Similarly,  two independent columns can be orthonormalized  to formulate a quaternion circle
\begin{equation}
\bm{q}=
     \cos\left(\chi\right) \bm{\alpha}_{orth_1}+
    \sin\left(\chi\right)\bm{\alpha}_{orth_2},
\quad \chi\in\left[-\pi,\pi\right]
\end{equation}

\section{Incompatible Case}
\begin{figure}[h]
	\centering
	\begin{tabular}{cccc}
		\includegraphics[width=0.2\linewidth]{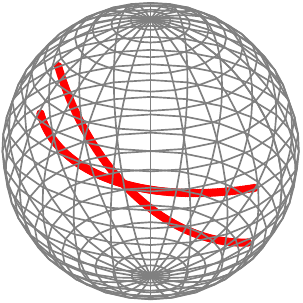}&
		\includegraphics[width=0.2\linewidth]{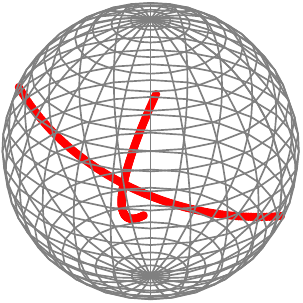}&
		\includegraphics[width=0.2\linewidth]{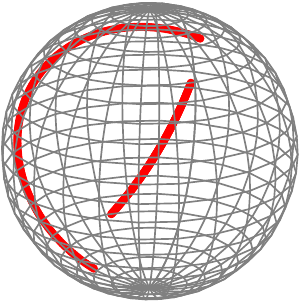}&
		\includegraphics[width=0.2\linewidth]{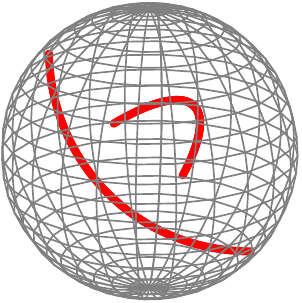}
		\\
		(a)&(b)&(c)&(d)
	\end{tabular}
	\caption{Illustration of incompatible case in $\mathbb{R}^3$. The red curves are the stereographic projection of half quaternion circle. (a) and (b) The quaternion circles intersect in $\mathbb{S}^3$. Therefore, after stereographic projection, they still intersect in $\mathbb{R}^3$. (c) and (d) The quaternion circles do not intersect in $\mathbb{S}^3$ and the corresponding curves also do not intersect in $\mathbb{R}^3$.}
	\label{fig:no-intersection}
\end{figure}

Typically, people believe that  two different correspondences can minimally solve the rotation estimation problem~\cite{parra2018guaranteed,peng2022arcs}. However, we  argue that two arbitrary inputs may not compute a compatible solution (see Fig.\ref{fig:no-intersection}). Note that in this case, we can still obtain a solution that minimize a loss function.

Without loss of generality, given $\{\bm{x}_1,\bm{x}_2\}$ and $\{\bm{y}_1,\bm{y}_2\}$, the target is to solve a rotation $\mathbf{R}\in\mathbb{SO}(3)$ that can meet $\mathbf{R}\bm{x}_1=\bm{y}_1$ and $\mathbf{R}\bm{x}_2=\bm{y}_2$. Traditionally, a  objective function in the least squares sense is established 
$
\min_{\mathbf{R}\in\mathbb{SO}(3)}\sum_{i=1}^{2}\|\mathbf{R}\bm{x}_i-\bm{y}_i\|^2
$, which just minimizes the loss and does not imply the solution can meet the given rotation constraints simultaneously.

In fact,
\begin{equation}
	\left\{
	\begin{aligned}
		\mathbf{R}\bm{x}_1=\bm{y}_1 \xrightarrow{Eq.~\eqref{eq:span}}\text{(Quaternion Circle 1)}:\bm{q}=\sin(\chi_1)\bm{\beta}_{1,1}+\cos(\chi_1)\bm{\beta}_{1,2}
		\\
		\mathbf{R}\bm{x}_2=\bm{y}_2 \xrightarrow{Eq.~\eqref{eq:span}}\text{(Quaternion Circle 2)}:\bm{q}=\sin(\chi_2)\bm{\beta}_{2,1}+\cos(\chi_2)\bm{\beta}_{2,2}
	\end{aligned}
	\right.
	\xrightarrow{\text{see ref. \cite{strang2012linear}}}
	\left\{
	\begin{aligned}
		\varnothing\cdots\textcircled1
		\\
		\pm\bm{q}^*\cdots\textcircled2
	\end{aligned}
	\right.
\end{equation}
Specifically, every rotation constraint responds to a Quaternion Circle, which is the intersection of a two-dimensional hyper plane (cross the origin point) and the unit quaternion sphere surface in $\mathbb{R}^4$. The intersection of two \textbf{different} quaternion circles can be considered using the intersection of two hyper plans to intersect the sphere surface. 

Note that the intersection of two \textbf{different} two-dimensional hyper planes (both cross the origin point) in $\mathbb{R}^4$ has two cases~\cite{strang2012linear}.
\begin{enumerate}
	\item Intersect only at the origin. In this case, the intersection of two hyper planes cannot intersect the sphere surface. Therefore, the two quaternion circles cannot intersect.
	\item Intersect in a straight line (cross the origin point). In this case, the intersecting line will meet the unit sphere surface at two symmetric points $\pm\bm{q}^*$, which are also the intersection points of two quaternion circles.
\end{enumerate}
\subsection{ Incompatible Example}
\begin{figure}[h]
	\centering
	\includegraphics[width=0.40\linewidth]{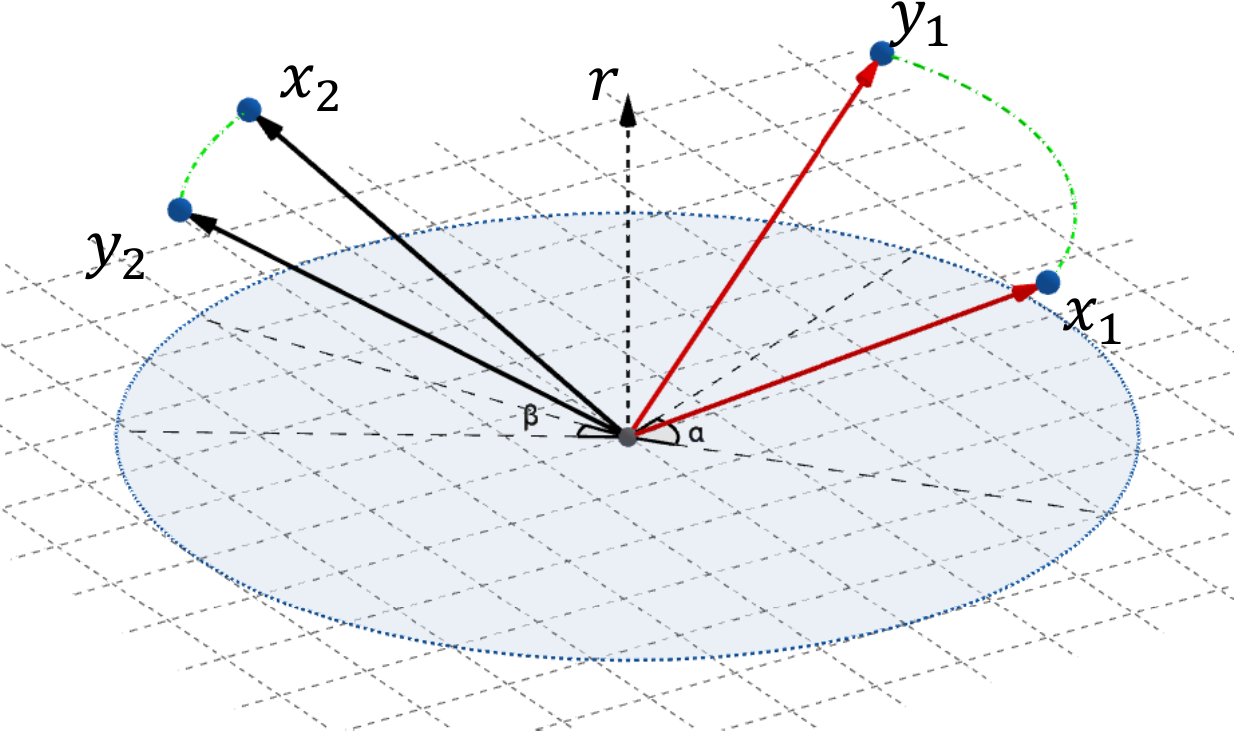}
	\caption{One incompatible example. Two input rotation constraints share the same rotation axis $\bm{r}$ and have different rotation angles.}
	\label{fig:incom}
\end{figure}

In this part, we show a visual example to demonstrate the incompatible case, and the setting is shown in Fig.~\ref{fig:incom}. Given two correspondences $\{\bm{x}_1,\bm{x}_2\}$ and $\{\bm{y}_1,\bm{y}_2\}$, and they share the same rotation axis $\bm{r}\in\mathbb{S}^2$; however, they have different rotation angle $\alpha$ and $\beta$.   In other words, $\bm{y}_1=\exp(\alpha[\bm{r}]_\times)$ and $\bm{y}_2=\exp(\beta[\bm{r}]_\times)$.

When $\bm{x}_1-\bm{y}_1$ and $\bm{x}_2-\bm{y}_2$ are uncorrelated, the rotation axis $\bm{r}$ can be determined uniquely by solving
\begin{equation}
	\left\{
	\begin{aligned}
		\bm{r}^T\left(\bm{x}_1-\bm{y}_1\right)=0
		\\
		\bm{r}^T\left(\bm{x}_2-\bm{y}_2\right)=0
	\end{aligned}
	\right.
\end{equation}
However, based on the calculated rotation axis $\bm{r}$, the rotation angles are different, which is a typical incompatible example. In this case, we cannot find a rotation that can meet the two given constraints simultaneously.

\end{document}